\RequirePackage{fix-cm}
\documentclass[twocolumn]{svjour3}          
\smartqed  
\usepackage{dsfont}
\usepackage{times}
\usepackage{epsfig}
\usepackage{epstopdf}
\usepackage{amsmath}
\usepackage{amssymb}
\usepackage{subfigure}
\usepackage{caption}
\usepackage{tabularx}
\usepackage[table]{xcolor}
 \usepackage{cite} 
 \usepackage[normalem]{ulem}
\usepackage[ruled,vlined,linesnumbered]{algorithm2e}
\usepackage[noend]{algpseudocode}
\usepackage[colorlinks,
            linkcolor=blue,
            anchorcolor=blue,
            citecolor=blue,
            bookmarks=false
            ]{hyperref}
\usepackage{booktabs}
\usepackage{multirow}
\usepackage[section]{placeins}
\makeatletter
\def\BState{\State\hskip-\ALG@thistlm}
\makeatother
\definecolor{myGreen}{HTML}{33FF00}
\definecolor{myRed}{HTML}{FF3030}
\definecolor{myGrey}{HTML}{AA5555}
\definecolor{myWhite}{HTML}{FFFFFF}
\definecolor{maroon}{cmyk}{0,0.87,0.68,0.32}
\definecolor{petr}{HTML}{5555FF}
\definecolor{josef}{HTML}{FF3030}

\journalname{International Journal of Computer Vision (IJCV)}
\begin{document}\sloppy

\title{A Review and A Robust Framework of Data-Efficient 3D Scene Parsing with Traditional/Learned 3D Descriptors }




\author{Kangcheng Liu}


\maketitle

\begin{abstract}
Existing state-of-the-art 3D point cloud understanding methods merely perform well in a fully supervised manner. To the best of our knowledge, there exists no unified framework that simultaneously solves the downstream high-level understanding tasks including both segmentation and detection, especially when labels are extremely limited. This work presents a general and simple framework to tackle point cloud understanding when labels are limited. The first contribution is that we have done extensive methodology comparisons of traditional and learned 3D descriptors for the task of weakly supervised 3D scene understanding, and validated that our adapted traditional PFH-based 3D descriptors show excellent generalization ability across different domains. The second contribution is that we proposed a learning-based region merging strategy based on the affinity provided by both the traditional/learned 3D descriptors and learned semantics. The merging process takes both low-level geometric and high-level semantic feature correlations into consideration.  Experimental results demonstrate that our framework has the best performance among the three most important weakly supervised point clouds understanding tasks including semantic segmentation, instance segmentation, and object detection even when very limited number of points are labeled. Our method, termed Region Merging 3D (RM3D), has superior performance on ScanNet data-efficient learning online benchmarks and other four large-scale 3D understanding benchmarks under various experimental settings, outperforming current arts by a margin for various 3D understanding tasks without complicated learning strategies such as active learning. 


\vspace{-2mm}
\end{abstract}

\vspace{-1mm}
\section{Introduction}
\label{intro}
\vspace{-2mm}
3D vision has great potentials in autonomous driving and robotics grasping \cite{liu2022industrialTIE, liu2022robustmm}. We tackle the 3D scene understanding problem, which typically consists of the three most important downstream tasks: 3D point cloud semantic segmentation, instance segmentation, and object detection.  It becomes increasingly important recently with the wide deployment of 3D sensors, such as LiDAR and RGB-D cameras. The 3D point clouds are the raw sensor data obtained by 3D sensors and the most common 3D data representation for scene understanding. The 3D data processing and scene understanding techniques have large potentials in the applications such as 3D robotic grasping, autonomous driving, and industrial applications \cite{liu2017avoiding}. \\
However, the 3D point cloud annotation often requires a long time and intensive manual labor. Besides, the majority of point cloud understanding methods rely on heavy annotations. For instance, it requires approximately half an hour per scene with thousands of scenes for ScanNet \cite{dai2017scannet} or S3DIS. Though existing point clouds understanding methods have achieved good results on these datasets, it is difficult to directly extend them to new scenes since they require a large number of high-quality labels at training; but not all the scenes contain a rich number of labels. For large-scale common indoor/outdoor scenes in robotics interaction and autonomous driving, it becomes more unrealistic. Therefore, weakly supervised learning (WSL)-based 3D point clouds understanding is highly in demand. Motivated by the success of WSL in images, many works start to tackle weakly supervised understanding with fewer labels, but great challenges remain. In general, the previous methods suffers from a lot of limitations. The graph network-based 3D WSL is proposed in \cite{wang2020weakly}. However, it still relies on heavy annotation costs for semantic labeling of 2D images projected from 3D point clouds, as well as the information loss when the 3D point clouds are transformed to 2D images, The MPRM has complicated pre-processing and pre-training process,  and the customized sub-clouds level labeling is also required, The recently proposed contrastive learning-based pretraining methods lack relationship mining both in low-level geometry and high-level semantics. Therefore, there is a lot of room to explore in how to fully unleash the capacity of WSL to make full use of weak-labeled 3D points in the limited annotation cases, and mining semantic/geometric correlations among the weakly-labeled regions and also the unlabeled regions. Also, how to leverage the both the 3D geometry information and 3D semantics to conduct representation learning in an end-to-end manner remains a problem to be solved. 
\begin{figure}[t!]
\setlength{\abovecaptionskip}{-0cm}
\setlength{\belowcaptionskip}{-0cm}
\centering
\includegraphics[scale=0.451]{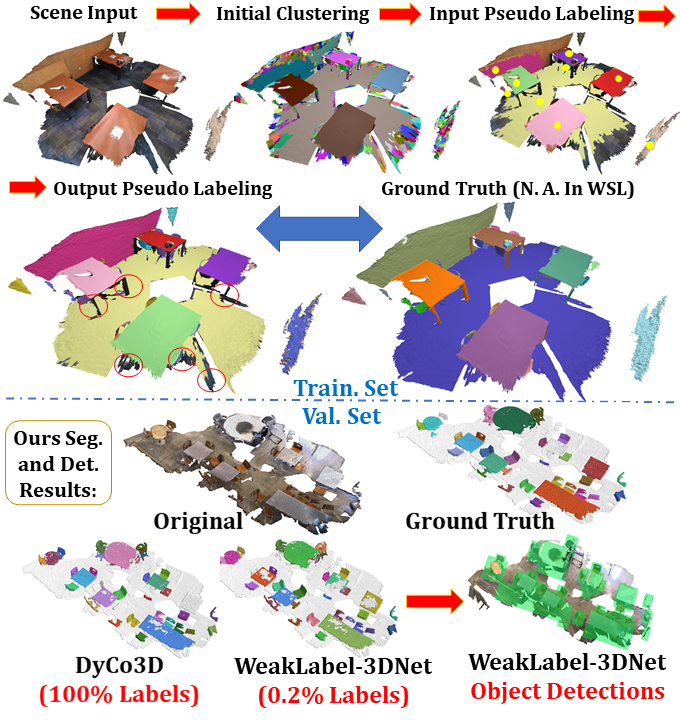}
\caption{Above the dash line shows our learning based pseudo label generation by the proposed learning-based region merging. The ground truth is not available in weakly supervised learning (WSL). The limited labeled points in weakly supervised learning are colored as golden yellow. Below shows our instance segmentation compared with current SOTA DyCo3D and object detection results respectively with merely 0.2\% labeled points.}
\label{Intro}
\vspace{-0.1000108cm}
\end{figure}

Motivated by challenges above in data-efficient 3D scene understanding, we study how to take advantage of low-level geometries to make full use of limited labels to realize multi-tasks point clouds WSL involving 3D semantic segmentation, 3D instance segmentation, and 3D object detection. As illustrated in Figure \ref{Intro}, we utilize an unsupervised method for generating initial regions by region expansion based on local normal, curvature, and traditional or learnt 3D descriptors, which encompasses prominent geometric characteristics of 3D objects. Next, we put forward to utilize the backbone network to produce region-level similarity prediction, which predicts the similarities among regions in the latent space. Then the region merging is applied iteratively to aggregate similar regions guided by both the traditional or learnt 3D descriptors and high-level semantic relationships to produce pseudo labels. We design self-supervised learning schemes to optimize the network with data augmentation losses to merge and propagate the weak labels to semantically similar regions. We directly use the results of instance segmentation to provide supervisions for object detection.
\vspace{-3mm}


To the best of our knowledge, our work is the first unified framework to tackle the weakly supervised multi-tasks 3D point clouds understanding. Our proposed framework attains nearly comparable segmentation performance with existing fully supervised state-of-the-arts (SOTAs) and significantly outperforms current weakly supervised SOTAs. 
In this work, we have largely extended the preliminary version of our works for the weakly supervised 3D scene understanding~\cite{liu2022weaklabel3d}, and have the following contributions:
\vspace{-0.51mm}
\begin{enumerate}
\item Firstly, we have done extensive methodology comparisons of traditional and learnt 3D descriptors for the task of weakly supervised 3D scene understanding. we have validated by experiments that our adapted PFH-based 3D descriptors show excellent adaptation and generalization ability across different domains.
\item
Secondly, we have proposed a learning-based region merging method based on the affinity provided by both the traditional/learnt 3D descriptors and the semantics. We have demonstrated by extensive experiments that both our proposed adapted traditional and simple contrastive-learning based learnt 3D descriptors can be integrated with our method to achieve weakly supervised 3D scene understanding with SOTAs performance and excellent rotational robustness. Also, it is demonstrated that our framework can generate high-quality pseudo labels compared with existing approaches.
\item We propose a data augmentation scheme to make the utmost use of weak labels by propagating them to similar points in latent space. And the effectiveness of JS divergence compared with the original mean error (ME) or mean squared error (MSE) loss function in the data augmentations loss is demonstrated.
\item State-of-the-art performance has been achieved by our framework with extensive experiments on publicly available ScanNet benchmarks and lots of other indoor/outdoor benchmarks including S3DIS, KITTI, and Waymo with diverse experimental circumstances. Our comprehensive results have provided baselines for future researches in 3D WSL.
\end{enumerate}
\vspace{-1mm}
\section{Related work} 
\subsection{Traditional versus Learnt 3D Descriptors}
\vspace{-2mm}
Extracting a discriminative local descriptor is very significant for downstream tasks of 3D scene understanding. In the past few years, various 3D descriptors have been proposed. The 3D local feature description is essentially extracting a feature vector around the query point to describe 3D local geometry. The 3D descriptor can be further divided into the histogram-based and signature-based approaches. The histogram-based approaches encode the local geometric variations and put them into the histogram. The typical histogram-based approaches include PFH and FPFH, and the typical signature-based approaches include SHOT. The differences between FPFH and PFH lie in following aspects. Firstly, FPFH merely has partial connected neighbours, while PFH has fully connected neighbours. And the range of neighbourhood is also different. Secondly, in PFH, each edge is counted only once, while in FPFH, a portion of edges are counted twice. Finally, for $N$ points each having $k$ points in the neighbourhood, the computational complexity of PFH is $O(Nk^2)$. And the FPFH has much less computational complexity, which is $O(Nk)$. According to our experiments in Section \ref{sec_experi}, the PFH demonstrates better 3D scene understanding performance compared with FPFH, although PFH is relatively computational intensive.
The concept of local reference frame has been proposed by SHOT to build a canonical pose of the local neighborhood. By this kind of design, it can achieve the rotational robustness and 6D-pose independence. Based on or similar to the above mentioned descriptors, many signature-based approaches such as the Heat Kernel Signature (HKS), and the Wave Kernel Signature (WKS), the Scale-Invariant Heat Kernel Signatures (SIKS) have been proposed, which are based on HKS. We have also analysed and compared the traditional visual similarity-based descriptor and the voxel cloud connectivity-based descriptor. The voxel connectivity-based approach can encourage generating segmented regions without crossing object boundaries by means of seeding methods based in 3D space and the flow-constrained local iterative clustering using color and geometrical features. The details of them and their advantages and drawbacks are illustrated in Subsection \ref{des_para}. Extensive evaluations and comparisons of these descriptors for various tasks of 3D scene understanding are shown in experimental results in Section \ref{sec_experi}. The detailed performance of these descriptors for the tasks of oversegmentation, semantic segmentation, and instance segmentation are all evaluated and compared in a detailed way. Moreover, we have proposed the adapted traditional and simple contrastive-learning based learnt 3D descriptors can be integrated with our method to achieve weakly supervised 3D scene understanding with SOTAs performance.

\vspace{-5mm}
\subsection{Learning-based Point Clouds Understanding Methods}
\vspace{-3mm}
Recently, various learning-based approaches have been proposed to tackle scene understanding in both 2D vision and 3D vision \cite{yuzhi2020legacy, liu2019deep}. Deep network-based approaches are widely adopted for point clouds understanding \cite{liu2020fg}. The fully supervised approaches can be roughly categorized into voxel-based \cite{liu2022weakly, liuws3d}, projection-based, and point-based methods \cite{liu2022fg, liu2021fg, liu2023fac}. Many recent works proposed to pre-train networks on source datasets with auxiliary tasks such as the low-level point cloud geometric registration \cite{xie2020pointcontrast}, the local structure prediction, the completion task of the occluded point clouds \cite{wang2021unsupervised}, and the high-level supervised point cloud semantic segmentation \cite{eckart2021self}, with effective learning strategies such as contrastive learning and generative models. Then, they fine-tuned the weights of the trained networks for the target 3D understanding tasks to boost performance on the target dataset. However, all the above methods require accessibility to high-quality fully annotated training data, which are hard to obtain for large-scale 3D scenes. It should be noted that 2D image and point cloud can be reciprocal in both scene understanding and generation. Recent approaches propose using distilled information from 2D image segmentation to assist 3D scene understanding. Also, it has recently be studied that the image can bridge the big semantic gap between the modalities of text and 3D shapes. Also, the LiDAR-based approaches are of significance to many industrial applications such as UAV/robotics inspections~\cite{liu2022robustmm, liu2022industrialTIE, liu2022semi} and robotic enhanced large-scale localization in the diverse complex environments~\cite{liu2022light, liu2022weaklabel3d, liu2022robustcyb, liu2022robust, liu2022integratedtrack, liu2023dlc, liu2022enhanced, liu2022enhancedarxiv, liu2022lightarxiv}, and large-scale robotic scene parsing~\cite{liu2021fg, liu2022fg, liu2020fg}, as well as robotic control as well as manipulation applications~\cite{liu2017avoiding, liu2023lidar, liu2022integrateduav, liu2022integratednoise, liu2022datasetsicca, liu2023learning}, etc. Differently, we make the first attempt in traditional and learnt 3D descriptor guided weakly supervised point cloud segmentation.

\vspace{-1.1mm}
\vspace{-2mm}

\subsection{Weakly Supervised methods for Point Clouds Understanding}

\vspace{-1mm}
The weakly supervised approaches for point cloud understanding are effective manners to reduce high annotation burdens \cite{liu2022semi}. Many preliminary attempts have been tried including labeling a small portion of points \cite{xu2020weakly, liu2021one, li2022hybridcr, hou2021exploring} or semantic classes \cite{wei2020multi}. Current approaches for weakly supervised 3D scene understanding can be divided into three main categories: consistency learning \cite{xu2020weakly, shi2021label}, pseudo label-based self-training \cite{liu2021one, cheng2021sspc}, and contrastive pre-training \cite{hou2021exploring, xie2020pointcontrast}. However, current weakly supervised point cloud understanding approaches are far from mature and have their own limitations. The graph-based 3D WSL was proposed to transform point clouds to images for obtaining semantic map, but image-level labels are required for training. Sub-cloud annotations \cite{wei2020multi} require the extra labour to separate sub-clouds and to label points within the sub-clouds.  Directly extending current art methods with weak labels for training will result in a great decline in performance \cite{liu2022fg} if label percentage drops to a certain value, which is less than 1\textperthousand. Self-training techniques have been utilized \cite{liu2021one} to design a two-stage training scheme to produce pseudo labels from weak labels with the 3D scenes, but it is only tested for the semantic segmentation task with limited performance. Xu et al. \cite{xu2020weakly} adopts semi-supervised training strategies combining training with coarse-grained scene class level information and with partial points using on tenth labels, but their test datasets are limited and it is tough to uniformly choose points to label. The network is elaborately made to approximate the gradient during the learning process, where the auxiliary 3D spatial constraints and color-level evenness were also considered in the network optimizations. However, the approach was restricted to the object part segmentation, and it is difficult to annotate points in a well-proportioned and homogeneous way as required. The unsupervised pre-training \cite{hou2021exploring} shows great capacity in unleashing the potential of weak labels to serve for complicated tasks, such as instance segmentation. But merely utilizing pre-training can not make full utilization of the weak labels, which results in dis-satisfactory performance. The concurrent work also explores the weakly-supervised video anomaly detection by magnitude contrastive learning \cite{MGFN} and the weakly supervised semantic segmentation with image-level supervision \cite{qi2016augmented}. However, our studied modality which is 3D point cloud is different in the modality and has essentially different properties with images or videos.

\subsection{3D Semantic/Instance Segmentation and Object Detection}
Recent studies have produced many elaborately designed networks for 3D semantic/instance segmentation \cite{jiang2020pointgroup} and object detection \cite{qi2019deep}. However, they all rely on full supervision. Recently, TWIST \cite{chu2022twist} also employs self-training-based approach to conduct effective semi-supervised learning. They innovatively proposed novel proposal re-correction module to filter out the low-quality proposals and enhance the pseudo label quality. In addition, many frameworks focus only on a single task, or two similar tasks \cite{pham2019jsis3d, wen2020cf}, and the relationships mining between those interconnected or complementary tasks, such as correlations between 3D instance segmentation and object detection, and the relationship between the 3D low-level geometry and high-level semantics, are rarely explored. 
\section{Proposed Methodology}
\label{sec_metho}
\vspace{-0.2mm}
We propose a general framework to tackle weakly supervised 3D understanding. Firstly, do to that various traditional or learnt descriptors can be integrated seamlessly with our proposed approach to conduct region merging, we give a comprehensive analysis of advantages and drawbacks of both traditional and learning-based descriptors in Subsection 3.1, and propose the adapted PFH-based descriptor with density robustness and an unsupervised contrastive learning based descriptor to achieve 3D scene understanding. The network backbone and the baseline framework for self-training are detailed in Subsection 3.2. The proposed region merging strategies for weakly supervised 3D scene understanding based on both traditional or learnt 3D descriptors and learnt semantics are given in Subsections 3.3 and 3.4 for semantic/instance segmentation and object detection, respectively.

\subsection{Traditional Versus Learning-based Descriptors}
In this Subsection, we first illustrate our adapted PFH-based descriptor. Then we illustrate other traditional descriptors, and discussed their advantages and disadvantages. Next, we have designed a simple contrastive learning-based descriptor that can achieve SOTAs performance in oversegmentation and downstream tasks. For the other learning-based descriptor, we have also detailed the advantages and disadvantages of them and done comprehensive comparisons in our experiments. 
\label{des_para}

\subsubsection{Our Proposed Adapted PFH-based Descriptors}

\label{subsection_pfh}
\begin{figure}[t]
\centering
\includegraphics[scale=0.43]{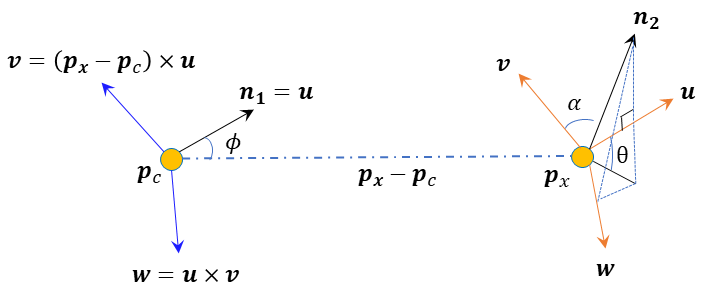}
\caption{Illustrations of adapted PFH-based 3D descriptor.}
\label{fig_des_pfh}
\vspace{-0.36 cm}
\end{figure}

In this Subsubsection, we illustrate our adapted PFH feature-based descriptors. We select the PFH-based \cite{rusu20113d} feature descriptor for its simplicity and robustness to 6D pose transformation \cite{rusu20113d}. Different from the original PFH-based \cite{rusu20113d} feature descriptor which utilizes the k-nearest neighbor of the point, we utilize the radius ball query to improve the robustness to the random noise. Also, we have discarded the point distances in the original PFH-based 3D feature descriptor for the fact that the point distances are easily influenced by the point density. The procedure of obtaining our adapted PFH-feature descriptor can be summarized as follows: for the select center point $p_c$, we find its neighbors with radius $r$. As shown in Figure \ref{fig_des_pfh}, denote $n_1$ as the surface normal at $p_c$, and $n_2$ as the surface normal at a neighbouring point $p_x$. Denote the relative position of two points on x axis, y axis, and z axis as $\Delta x, \Delta y, \Delta z$, then each pair of points gives a point set $[\alpha, \phi, \theta, \beta_1, \beta_2, \beta_3, d]$ given as follows:

\begin{scriptsize}
\begin{equation}
\begin{aligned}
\centering
    &\alpha= v \cdot n_2, \quad \phi = u \cdot \frac{p_x-p_c}{\|p_x-p_c\|_2},  \\&
    \theta=arctan (w \cdot n_2, u \cdot n_2), \quad d=\|p_x-p_c\|_2 ,  \\&
    \beta_1=\frac{\Delta x}{d}, \quad \beta_2=\frac{\Delta y}{d}, \quad \beta_3=\frac{\Delta z}{d}.
\end{aligned}
\end{equation}

\end{scriptsize}

\begin{figure}[t]

\centering
\includegraphics[scale=0.2536]{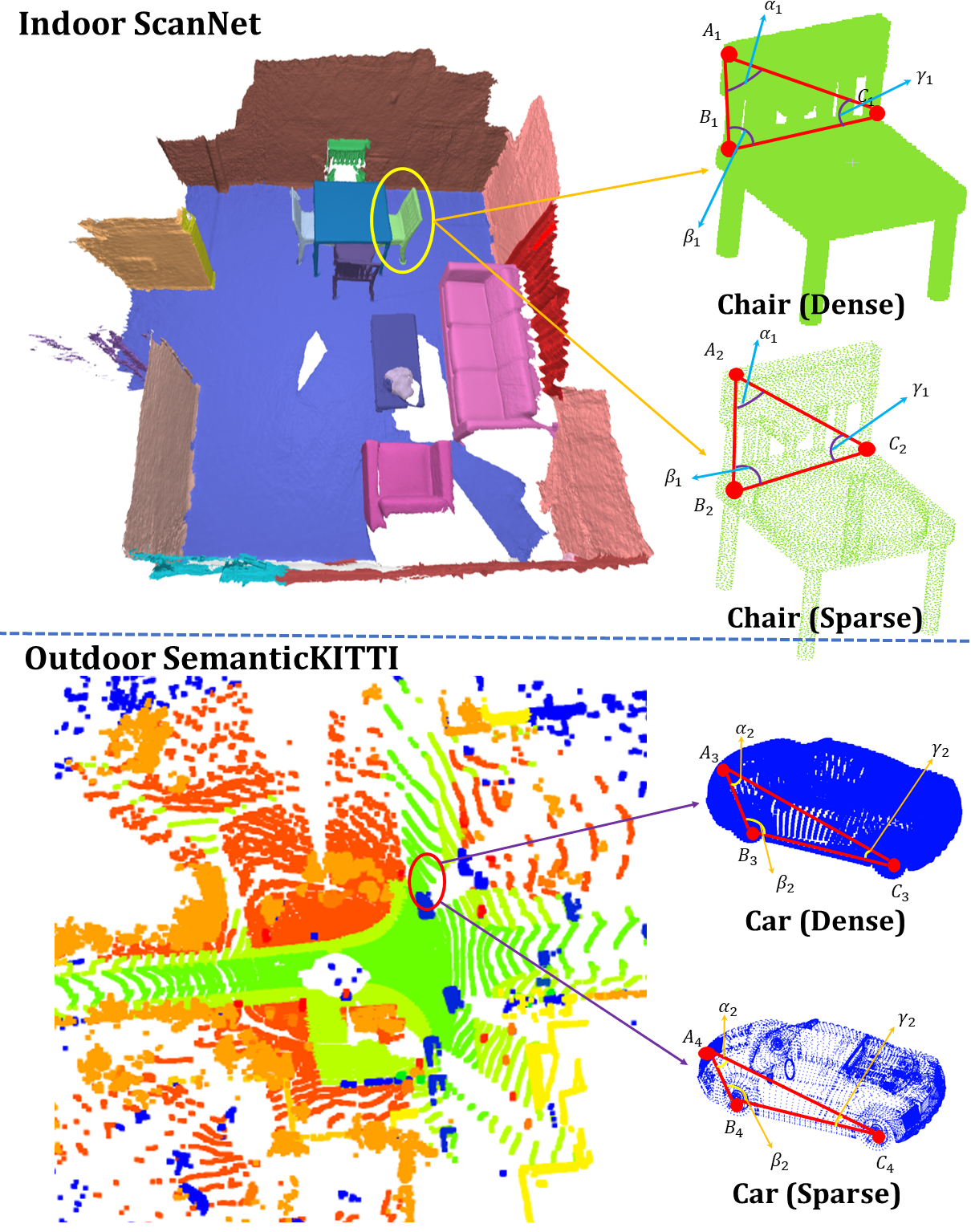}
\caption{The illustration of the robustness of relative angles to the point density.}
\label{fig_angle_pfh}
\vspace{-0.1cm}
\end{figure}

The original PFH \cite{rusu20113d} takes the set $[\alpha, \phi, \theta, d]$ to formulate the feature descriptor. It is very easily influenced by the point density especially in the outdoor scenes with sparse LiDAR points. The reason is that it takes the distances between points into consideration, which is sensitive to density for downstream tasks of 3D scene understanding. For example, for the outdoor case, the close-to-center point clouds will be denser, and the distance between point clouds will be smaller. Therefore, for the same object, it will often result in different descriptors because of the different densities. As shown in Figure \ref{fig_angle_pfh}, the relative angle is very robust to point density because the relative angle will remain the same if the density or scale of the point clouds changes. We have illustrated the density robustness in Figure \ref{fig_angle_pfh}. And for both the indoor and outdoor scenes, the relative angles will not change with respect to the point density. More specifically, as shown in Figure \ref{fig_angle_pfh}, we denote the same point on the geometry for the dense point clouds chair and the sparse point clouds chair as $A_1$ and $A_2$, and the same goes for other points such as $B_1$ and $B_2$, et al. The triangle $A_1B_1C_1$ is always identically equal to the triangle $A_2B_2C_2$. Therefore, the relative angle between points, such as $\alpha_1$,  $\beta_1$, and $\gamma_1$ will not change with the change of the point density. In our adapted PFH-based feature descriptor, the point distance in the original PFH-based feature representation is discarded. And our adapted PFH-based descriptor takes the set $[\alpha, \phi, \theta]$ to formulate the 3D feature descriptor. Finally, it is demonstrated by our extensive experiments that our adapted PFH-based feature descriptor has robust performance for both high-density indoor RGB scans such as S3DIS and ScanNet \cite{dai2017scannet}, and for low-density outdoor LiDAR scans such as SemanticKITTI \cite{behley2019semantickitti}. It is also demonstrated by experiments that our framework has good rotation robustness. 

Our adapted PFH-based feature is robust to random noise because the random noise merely results in the change in point distance. And the point distance is not considered in our adapted PFH-based feature. Also, according to our experiments, the random noise existing in our tested datasets will not have great influence on the final point clouds understanding for the fact that the noise model depends on the sensor. The same RGB-D camera or LiDAR have the same noise model. Also, in our setting for all datasets, the noise model remains the same in the training set and the testing set. Also, due to that our proposed RM3D is a representation learning-based model, the noise robustness of our approach is better as compared with traditional approaches. Also, it is demonstrated extensively by recent works in point clouds learning that representation learning-based model is very robust to noises and corruptions \cite{ren2022benchmarking}, \cite{ye2021learning}, \cite{liu2019relation}. Finally, the scene understanding performance of whole framework is not much influenced by noisy point clouds input such as in SemanticKITTI \cite{behley2019semantickitti} according to our experiments. 

\subsubsection{Other Traditional Descriptors}
In this Subsection, we do a comprehensive review of the traditional descriptors. We have also done experiments with many other traditional descriptors. The short illustrations of various of descriptors are summarized as follows.
\label{other_des}
\\
\textbf{SHOT}
 The Signature of Histogram (SHOT) \cite{salti2014shot} \cite{tombari2010unique} is also a popular 3D descriptor. The procedure of obtaining the SHOT descriptor can be summarized as follows: First of all, we need to divide the space into several small volumes. Secondly, we need to compute the local histogram of each volume. Thirdly, we need to concatenate the local histograms into a "signature". It should be noted that with the local reference frame (LRF) \cite{salti2014shot}, the signature is 6D pose invariant. Finally, we should normalize the "signature" into the sum of one. The major drawback of SHOT \cite{salti2014shot} 3D descriptor is the boundary effect. The boundary effect is that points at the edge of each volume should contribute to the neighbouring volume as well. And small perturbations of local reference frame change all the local histograms.   \\
\textbf{FPFH}
We have also done experiments with the FPFH. There are some major differences between the FPFH and PFH. Firstly, the FPFH descriptor has partial connected neighbors, while the PFH descriptor has fully connected neighbors. And the FPFH has partial connected neighbours. Also, denote the query radius as $r$ the FPFH has neighborhood range of $[r, 2r]$ while the PFH has the neighborhood range of $[0, r]$. In the PFH feature, each edge is counted once; while in the FPFH feature, some edges are counted twice. The FPFH has the computational complexity of $O(nk)$, while the PFH has the computational complexity of $O(nk^2)$. According to our experiments, the final performance of FPFH is a little inferior compared with PFH although it has a lower computational complexity.
\\
\textbf{The Heat Kernel Signature (HKS)}
Similar to PFH, the HKS \cite{sun2009concise} is also a point signature based on the properties of the heat diffusion process on a 3D shape. The HKS is an extension of the well known heat kernel to the temporal domain. The biggest advantage is that HKS can capture information about the neighbourhood of a point in a multi-scale way. While the disadvantage of it is also the high-computational cost. \\
\textbf{The wave kernel signature (WKS)}
The wave kernel signature \cite{aubry2011wave} and the scale invariant heat kernel signature (SIKS) \cite{bronstein2010scale} are all improved versions of HKS. The wave kernel signature is very distinctive for the fact that it represents the average probability of the quantum particle appearing at a specific location. By varying energy, WKS separates information from different Laplace frequencies. It is demonstrated by our experiments that WKS can also be integrated seamlessly with our proposed learning-based WSL 3D scene understanding framework. Also, The WKS \cite{aubry2011wave} is invariant to isometries and very robust to small non-isometric deformations compared with HKS \cite{sun2009concise}. \\
\noindent \textbf{The scale invariant heat kernel signature (SIKS)}
The SIKS \cite{bronstein2010scale} is also proposed for the non-rigid shape recognition. The biggest advantage of SIKS is that it can maintain the invariance under various transformations the shape experienced. It is very robust to various of transformations such as the isometric deformations, the missing of data, the topological noises, and also the global or local scale change. Also, the biggest disadvantage is that the computation is very slow trained with this kind of descriptor.  \\
\textbf{Visual Similarity-based 3D Model Retrieval}
The visual similarity-based 3D model retrieval \cite{chen2003visual} was also proposed. The main ideas are that if the two 3D models are similar, they also look similar in different viewing angles. However, as this work relies on the image-level description at multiple viewing angles, it is not very robust to the view angle selection and rotation of the target object.  \\
\textbf{Persistent/Point Feature Histograms (PFH/PSH)}
{The PSH was also proposed in \cite{rusu2008aligning} and is essentially the same as the PFH feature. As mentioned previously, according our experiments, the PFH relies on the point cloud distances, which is not robust to the 3D point density. Therefore, we also propose our adapted PFH feature, which is very robust to the low-density scenarios according to our experiments.}\\
\noindent
\textbf{Geometric Partition with Global Energy}
Also, the cut pursuit \cite{landrieu2017cut} \cite{landrieu2018large} is also proposed to produce the super-points based on the graph-cut algorithm.
It can be performed in an unsupervised manner to provide the segmentation results that are adaptive to the local geometrical complexity. For example, the regions obtained can be large simple shapes such as road and buildings, and can also be small shapes such as parts of cars and pedestrians.

\begin{figure}[t!]
\centering
\includegraphics[width=\linewidth]{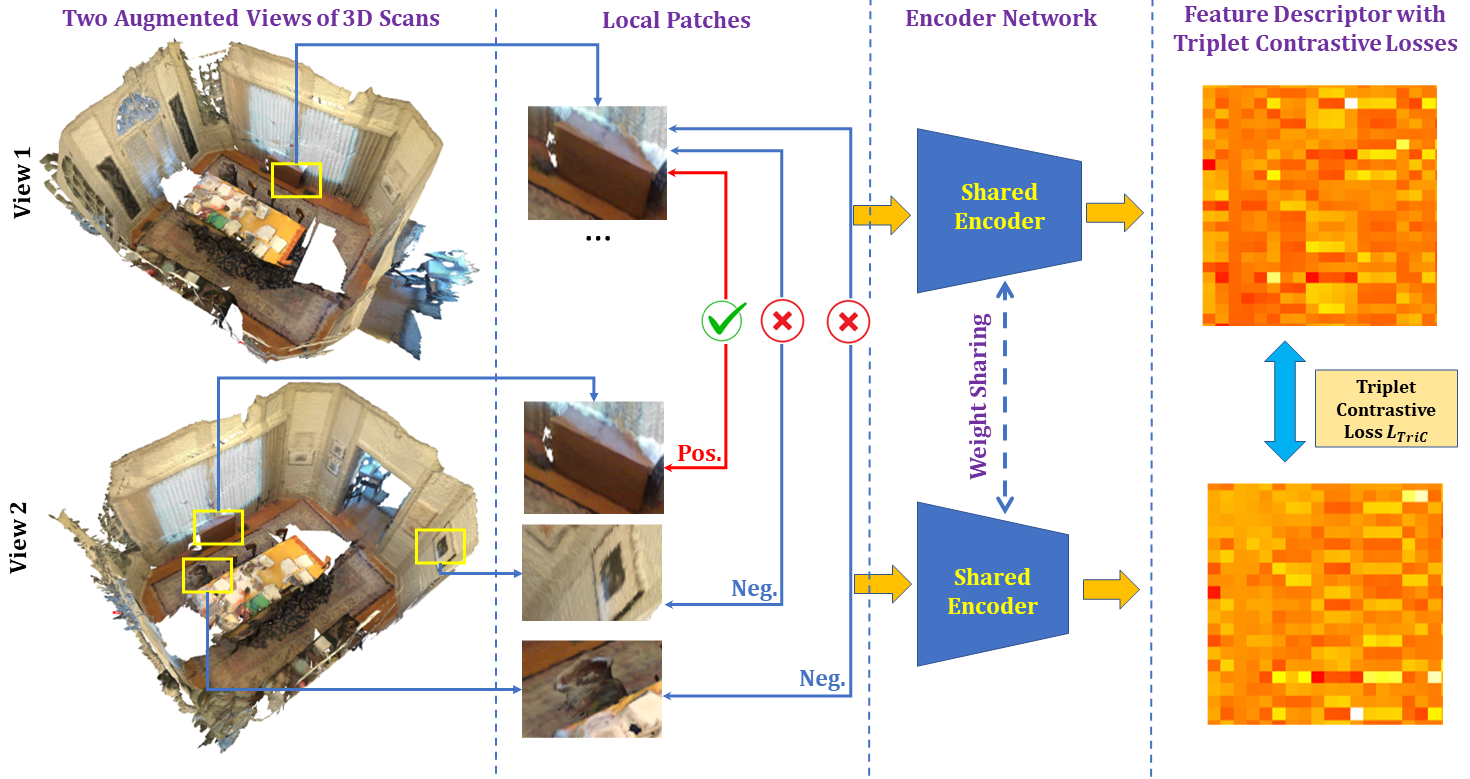}
\caption{The illustration of learning-based 3D descriptor.}
\label{fig_des_learn}
\vspace{-0.2cm}
\end{figure}

\begin{figure*}[t!]
\centering
\includegraphics[scale=0.140988]{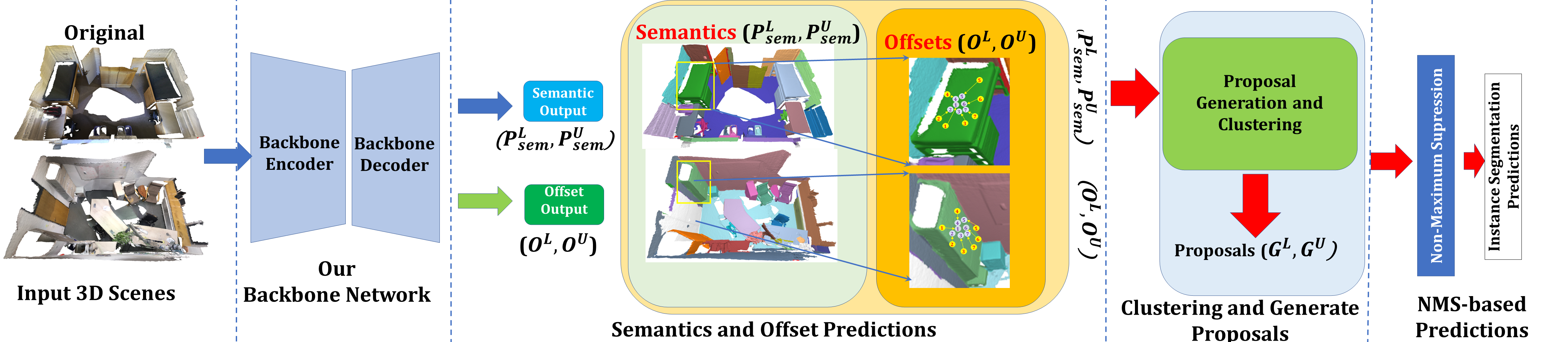}
\caption{The overview of the self-training-based instance segmentation framework for our proposed RM3D.}
\label{fig_ins_seg_frame}
\vspace{-3.29mm}
\end{figure*}

\subsubsection{Unsupervised Contrastive Learning-based Descriptor}
\label{subsection_des}

Recently, many learning-based 3D point cloud descriptors have been proposed and have demonstrated their performance in low-level registration \cite{tang2022multi} and high-level understanding \cite{jiang2018pointsift}. In this work, we propose a simple but effective learning-based descriptor which can outperform traditional descriptors and achieve superior performance in a single dataset without any transfer learning. Note that according to our experiments, although the learning-based descriptor can realize superior performance on a single dataset, the transfer learning performance is not that good, which also demonstrate the generalization capacity of current learning-based descriptors should be improved. And the traditional descriptors have the advantage of great generalization capacity across domains . It is demonstrated in Table \ref{table_transfer_learning_i} the transfer learning performance of our adapted PFH is better compared with Predator \cite{huang2021predator} in the indoor transfer learning between S3DIS and ScanNet. Next, we will introduce our contrastive learning-based descriptor, which can be integrated seamlessly to our weakly-supervised point clouds segmentation and detection framework because of its unsupervised characteristic.


As shown in Figure \ref{fig_des_learn}, to facilitate contrastive learning, we first break down the 3D scans into local patches and define the positive local pairs and negative local pairs. The positive pairs are defined as the pairs of the same physical place (i.e. the minimal physical point distance between the two point sets is less than 0.05m), while the negative pairs are defined as the pairs of different places (the two point sets that are at least 1m apart). For the number of positive and negative samples, we have also selected carefully. We select the point-based SparseConv \cite{graham20183d} as our backbone network.


For the learning of the local geometry, we have leveraged two loss functions, the original contrastive loss $L_{C}$ and the triplet $L_{TriC}$ contrastive loss for the network optimizations. Denote $d_{ij}$ as the distance between pair of points, and denote $y_{ij}$ as the ground truth label of the positive or the negative pairs. Denote the anchor for comparisons for contrastive learning as $S_a$ \cite{chen2020simple}, the positive sample for contrastive learning as $S_p$, and the negative sample for contrastive learning as $S_n$. Denote $N_{c}$ as the number of total samples, the two losses $L_{C}$ and $L_{TriC}$ are formulated as:

\begin{scriptsize}
\begin{equation}
\begin{aligned}
\centering
     L_{C}=\frac{1}{N_{c}}\sum_{n=1}^{N_{c}} y_{ij}d^2_{ij} + (1-y_{ij}) max (\tau -d_{ij}, 0)^2,
\end{aligned}
\label{eq_1}
\vspace{-6mm}
\end{equation}

\begin{equation}
\begin{aligned}
\centering
     L_{TriC}=\frac{1}{N_{c}}\sum_{n=1}^{N_{c}} max(d_{ij}(S_a, S_p)-d_{ij}(S_a, S_n)+\rho, 0).
\end{aligned}
\label{eq_2}
\vspace{-2mm}
\end{equation}
\end{scriptsize}

The $\tau$ and $\rho$ are thresholds. For the optimization in contrastive learning, we have leveraged triplet contrastive loss instead of the original contrastive loss for the fact that the original contrastive loss is too greedy. It merely aims at minimizing the distances between positive samples and maximizing the distance between negative samples. Therefore, the original contrastive loss can be easily trapped into a local minimum during network training. According to our experimental results, our proposed simple contrastive learning-based descriptor can be integrated into our framework with SOTAs 3D scene understanding performance.  
 \\
After the network is trained, the network gives a distinctive descriptor to describe different local 3D point cloud patches. The 3D local descriptor also serves for the following cluster-level region merging based on local similarities.
\vspace{-2mm}
\subsubsection{Other Learning-based Descriptors}
We have also done experiments with many other learning-based descriptors. For example, the Point-SIFT \cite{jiang2018pointsift} was proposed using directional encoding and a scale-awareness network to embed the local feature of the point scan. Therefore, the scale and the directional awareness can be largely improved. Also, the cut pursuit \cite{landrieu2017cut} \cite{landrieu2018large} is also proposed to produce the super-points based on the greedy graph-cut algorithm.  Then, they did some improvements using the deep neural network to obtain the feature embedding of the point clouds and combined them with the graph-structured deep metric learning to over-segment the point cloud \cite{landrieu2019point}. We have also done a comprehensive comparison with those learning-based descriptors and tested their generalization capacity. In the recent work Predator \cite{huang2021predator}, an overlapping attention block for early information exchange between the underlying codes of two point clouds is designed. In this way, the model is able to decode the latent representation into the characteristics of each point to predict which points are prominent and located in the overlapping region of the two point clouds in terms of the respective other point cloud. And the subsequent work \textit{Multi-Ins-Reg} \cite{tang2022multi} has extended the Predator to the registration of multiple point clouds with the proposed clustering algorithms to the correspondence between different clouds.



\vspace{0.5cm}

\begin{figure*}[t]
\centering
\includegraphics[width=\linewidth]{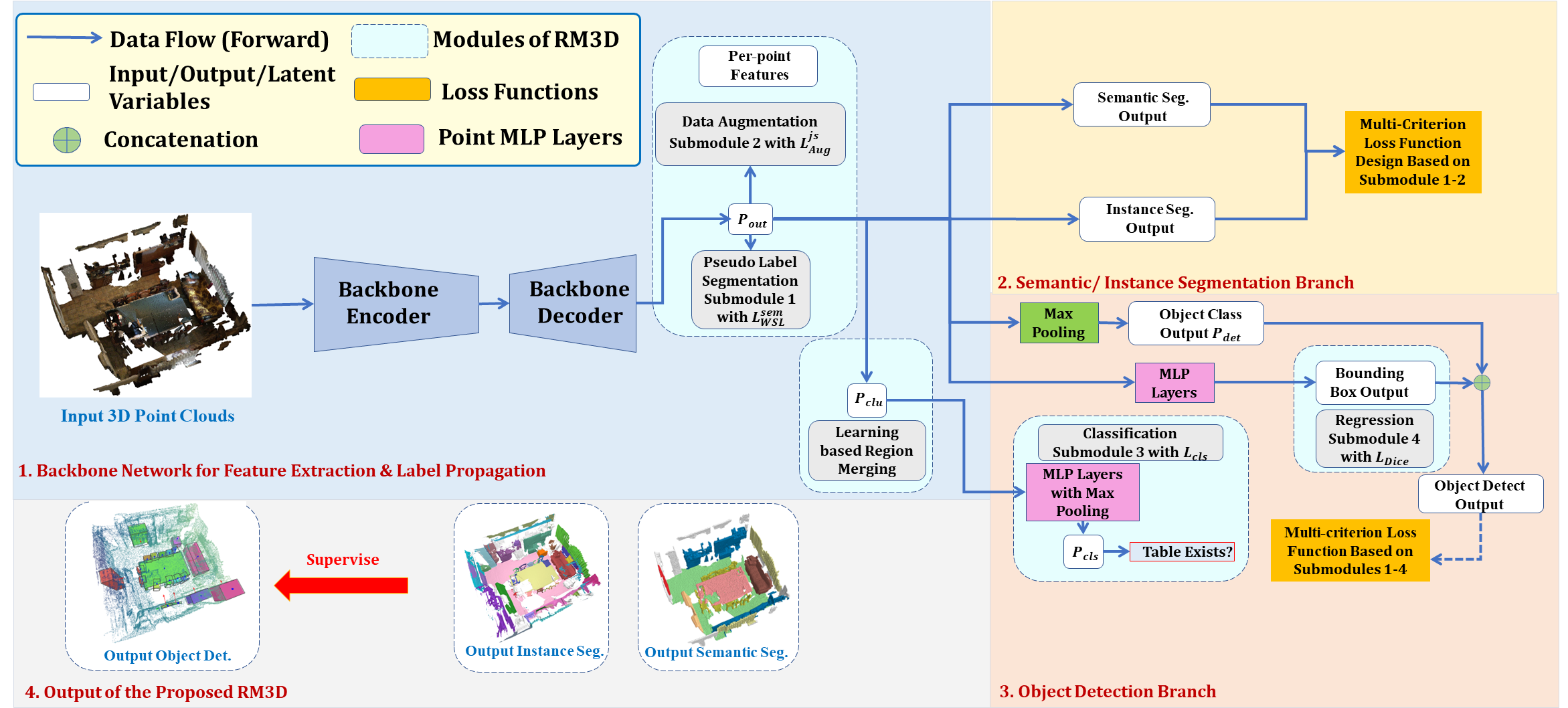}
\caption{\textbf{RM3D Architecture Overview}. It consists three components: 1. The backbone network adopts the same encoder-decoder structure to obtain the per-point features. 2. The semantic/instance segmentation branch. 3. The object detection branch are the two main output branches supervised by our proposed network optimization modules. We have unified the network backbone as SparseConv \cite{graham20183d} for various 3D scene understanding tasks including semantic segmentation, instance segmentation, and object detection.}
\label{fig_backbone}
\vspace{-3.06cm}
\end{figure*}

\subsection{Revisit Baseline Framework of Self-Training}

As our weakly supervised learning framework relies on the self-training, we detail the process of self-training for the tasks of semantic segmentation and instance segmentation respectively in this Subsection.  For the task of both semantic segmentation and instance segmentation, we adopt the backbone of SparseConv \cite{graham20183d}.
\subsubsection{Baseline for self-training-based instance segmentation}
\label{baseline_insseg}

  Firstly, in this Subsection, we detail the procedure of self-training-based instance segmentation. The instance segmentation branch is different from the semantic segmentation branch for the fact that clustering should be done to segment each instance and the offset of each point should be learnt. In this work, we develop a weakly supervised framework to do instance segmentation. Our framework is generally similar to the fully supervised instance segmentation framework PointGroup. However, we have proposed several designs to make the original framework suitable for weakly supervised instance segmentation in the limited annotation setting. Different from the previous self-training, we use confidence threshold to ensure these high-confidence regions are given pseudo labels during region merging with self-training. The procedures are shown in Figure \ref{fig_ins_seg_frame} as detailed as follows.

  Denote the $P_L$ as the labeled points and $P_U$ as the unlabeled points, and we denote $P^L_{sem}, O^L, G^L$ and $P^U_{sem}, O^U, G^U$ the point-level semantic predictions, the point-level offset, and the predicted instance proposal on $P_L$ and $P_U$, respectively. 

    We leverage the self-training pipeline to leverage the pseudo labels of the points in merged regions in the unlabeled data. The pipeline is summarized as follows:


The \textbf{first step} is the region merging stage. At the first iteration, the regions with pseudo labels propagate labels to the high-confidence similar regions utilizing our proposed learning-based region merging Submodule in the next Subsection. Then the pseudo label is updated and merged. Note that these high-confidence similar regions are regarded as labeled regions permanently in training. Therefore, the confidence is very significant because it ensures the regions can be merged are not only the similar ones but also the high-confidence predicted regions. The produced pseudo semantic label at the current iteration is $S^{U}_t$, and the predicted offset pseudo label is $O^{U}_t$. At the current iteration of the self-training, we utilize the pseudo labels $O^{U}_t$ and $S^{U}_t$ to guide the self-training process.

The \textbf{second step} is the training epoch for updating the current models. At the current iteration of the self-training, we utilize the updated pseudo labels to train and refine the network. We utilize the pseudo semantic labels $S^{U}_{(t)}$ and the pseudo offset labels $O^{U}_{(t)}$ to guide the network training at the current iteration and give the produced semantic prediction $S^{U}_{(t+1)}$ and offset vector $O^{U}_{(t+1)}$:
\begin{scriptsize}

\begin{equation}
\begin{aligned}
\centering
     L^{Ins}_{U}=L_{s}(S^{U}_t, S^{U}_{(t+1)})+ L_{o} (O^{U}_t, O^{U}_{(t+1)}).
\end{aligned}
\end{equation}

\end{scriptsize}

And similarly, for the labeled data,

\begin{scriptsize}

\begin{equation}
\begin{aligned}
\centering
     L^{Ins}_{L}=L_{s}(S^{L}_t, S^{L}_{(t+1)})+ L_{o} (O^{L}_t, O^{L}_{(t+1)}).
\end{aligned}
\end{equation}

\end{scriptsize}


The $L_{s}$ is the cross entropy loss on the semantic predictions $S^{u}$, and the $L_{o}$ is the regression term for regularizing both the direction and the $L_1$ distance of the predicted offset vector $O^{L}$. For a point cloud of $N_s$ points, the voting center regression loss can be formulated as:

\begin{scriptsize}
\begin{equation}
 \small
     L_{o}(O^{U}_t, O^{U}_{(t+1)})= \frac{1}{N_s} \sum_{i=1}^{N_s}(\|o^U_{t}-o^U_{t+1}\|-\frac{o^U_{t}}{\|o^U_{t}\|_2}\cdot \frac{o^U_{t+1}}{\|o^U_{t+1}\|_2}).
     \label{eq_instance}
\end{equation}
\end{scriptsize}
The summed training objective $L^{WSL}_{Ins}$ for weakly-supervised instance segmentation can be summarized as:

\begin{scriptsize}
\begin{equation}
\begin{aligned}
\centering
     L^{WSL}_{Ins}= L^{Ins}_{L} +  L^{Ins}_{U}.
\end{aligned}
\end{equation}
\end{scriptsize}


 Note that different from previous typical self-training, our proposed  self-training process merely uses our propagated labels, our model can use a more high-quality propagated label to refine the quality of final segmentation results, thus improving the performance of the whole framework.

As shown in Figure \ref{fig_backbone}, we have adopted a typical encoder-decoder-based network backbone to extract the per-point features. For the task of both semantic and instance segmentation, We have unified our backbone network to the Sparse Convolutional network (SparseConv).  The backbone network for the object detection is also SparseConv \cite{graham20183d}, and the strategy for doing object detection is based on VoteNet. The axis tightly aligned bounding box of each instance provided by the results of instance segmentation is selected as the initialization of pseudo ground truth bounding boxes for object detection. As shown in Figure \ref{fig_backbone}, we have utilized the network to produce the region-level similarity prediction. The proposed region-level similarity prediction strategy is detailed in the next Subsection \ref{sub_sec_merge}.

It should be noted that in the weakly supervised setting, we can not adopt the ScoreNet proposed in PointGroup because we can not localize the instance with limited annotations. Therefore, we simply average the semantic prediction scores of the point clouds belonging to the same instance. It should be noted that the clustering is merely required in the test time. During the test time, we cluster the points that are shifted by the learnt direction offset and distance vector offset using the width first search algorithm adopted by the PointGroup. It should be noted that the clustering merely happens in the test time. During the test time, we cluster the points that are shifted by the learnt direction offset and distance vector offset using the width first search clustering algorithm adopted by the PointGroup.  The directional and distance vectors are learnt by the voting center loss shown in Equation \ref{eq_instance} during the self-training.

\vspace{-1mm}
\subsubsection{Baseline for self-training-based semantic segmentation}
For the task of semantic segmentation, we adopt the backbone of SparseConv \cite{graham20183d}, which is the same as the backbone for instance segmentation. The difference exists in that the optimization functions for semantic segmentation are simpler. The training in semantic segmentation also adopts the self-training pipeline. The same as the instance segmentation task, the training objective $L^{WSL}_{Sem}$ is summarized as the sum of the loss for the labeled data $L^{Sem}_{L}$ and the loss for the unlabeled data $L^{Sem}_{U}$:

\begin{scriptsize}
\begin{equation}
\begin{aligned}
\centering
     L^{WSL}_{Sem}= L^{Sem}_{L} +  L^{Sem}_{U}.
\end{aligned}
\end{equation}
\end{scriptsize}

The same as before, the $L_s$ is the cross entropy loss for the semantic segmentation. And the summed semantic segmentation loss for labeled data and unlabeled data can be summarized as:
\begin{scriptsize}
\begin{equation}
\begin{aligned}
\centering
     L^{Sem}_{L}=L^{Sem}_{s}(S^{L}_t, S^{L}_{(t+1)}),
\end{aligned}
\vspace{-3mm}
\end{equation}
\end{scriptsize}

\begin{scriptsize}
\begin{equation}
\begin{aligned}
\centering
     L^{Sem}_{U}=L^{Sem}_{s}(S^{U}_t, S^{U}_{(t+1)}).
\end{aligned}
\end{equation}
\end{scriptsize}

The summed training objective $L^{WSL}_{Sem}$ can be summarized as:

\begin{scriptsize}
\begin{equation}
\begin{aligned}
\centering
     L^{WSL}_{Sem}= L^{Sem}_{L} +  L^{Sem}_{U}.
\end{aligned}
\end{equation}
\end{scriptsize}


The task of semantic segmentation only requires the prediction of the per-point semantics. Therefore, we combine the  $ L^{WSL}_{Sem}$ with the data augmentation loss $L^{js}_{Aug}$ proposed in the next Subsection \ref{sub_sec_merge} for the end-to-end training of the network. For the task of instance segmentation, we combine the  $ L^{WSL}_{Ins}$ with the data augmentation loss $L^{js}_{Aug}$ proposed in the next Subsection \ref{sub_sec_merge} for the end-to-end training of the network.

\subsection{Proposed Region Merging Strategies for Segmentation}
\label{sub_sec_merge}
\begin{figure}[t!]
\centering
\includegraphics[scale=0.180]{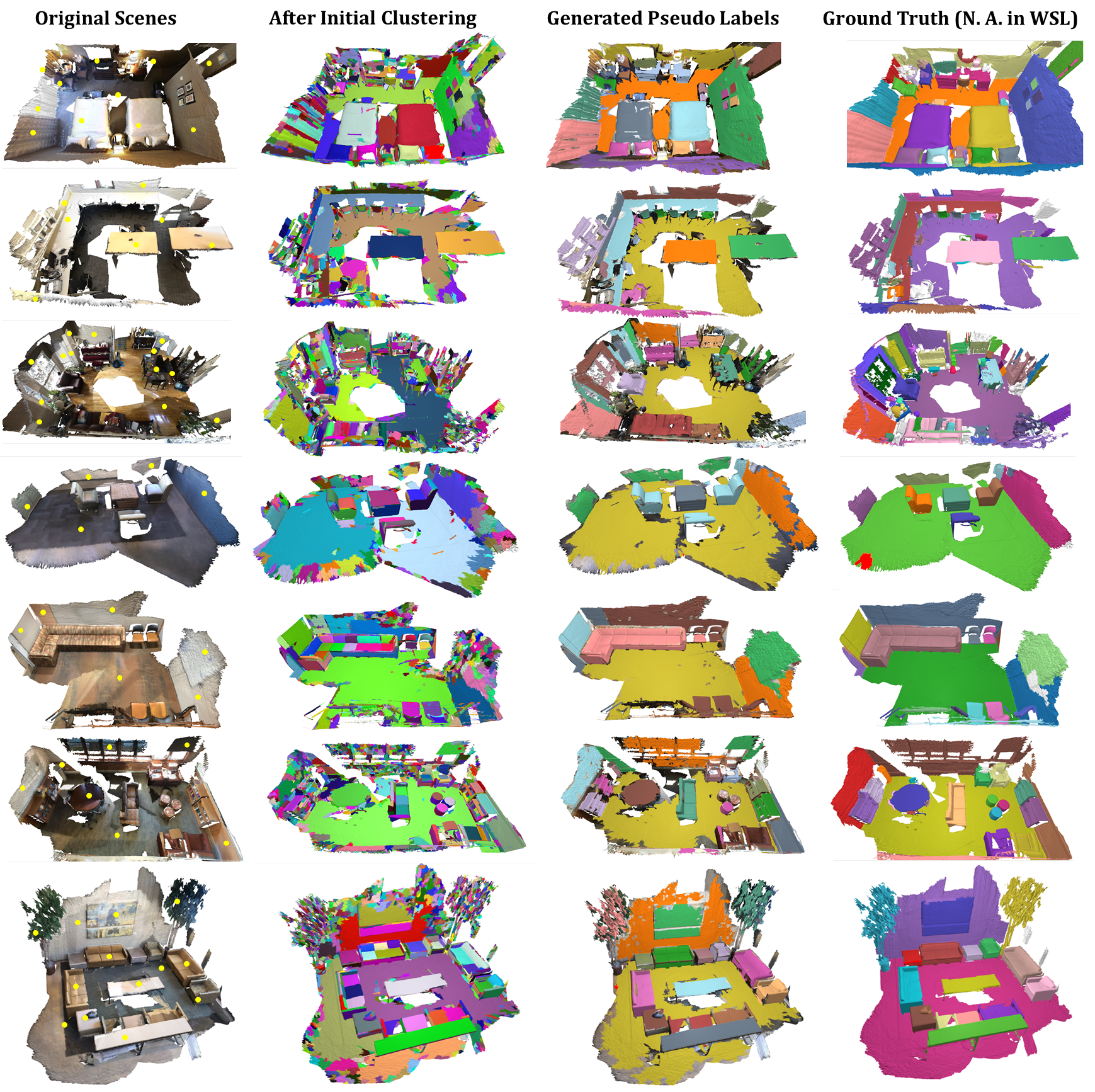}
\caption{Region expansion-based labeling results and comparisons with ground truth on ScanNet validation set.  The first column shows the original scene. Truly labeled points are indicated by yellow in the first column of original scene.  The second column shows the initial clustering results after PFH-based oversegmentation. The third column shows the final generated pseudo labels after self-training. The final column shows the ground truth instance segmentation results, which are not available in weakly supervised learning. The key finding is that our generated pseudo labels are very similar to the ground truth labels. Therefore, we can use our generated pseudo label to substitute the true label in the self-training for weakly supervised segmentation.}
\label{Fig_Region_Expansion}
\vspace{-0.1mm}
\label{Clust}
\end{figure}

Our proposed region merging strategies for segmentation is composed of two parts. The first part is the region-level similarity prediction strategy, and the second part is the learning-based region merging. The two parts are detailed as follows: \\
\textbf{The Region-Level Similarity Prediction Strategy} 
We have proposed a learning-based region merging method based on the affinity provided by both the traditional and learning-based 3D descriptors. We first use the PFH-based oversegmentation to obtain regions. The oversegmentation procedure is provided in the Appendix. After the oversegmentation, we can obtain the initial geometrically separated regions. Some randomly selected oversegmentation results are shown in the second column of Figure \ref{Fig_Region_Expansion}. It is apparent that our adapted PFH-based oversegmentation results can automatically divide the whole point clouds scene into geometrically well-separated regions, which are indicated by different colors. After the point clouds oversegmentation, we also obtain the original pseudo labels by expanding each labeled point to all the points in the region that includes the labeled point. For the regions containing more than one label point, we directly use the semantic of the maximum number of points as the inital pseudo label for the region. If two classes have the same number of labeled points, we have the priority for assigning the class that has larger number of points in the training set. Most importantly, in this Subsection, we propose an end-to-end approach to incorporate the traditional or learnt 3D local descriptors mentioned above for learning-based region merging.

To be more specific, we compute the normal vector of every individual region based on the widely adopted PCA analysis of local neighbouring points. Denote the average normal vector of a certain region as $\textbf{n}=(n_x, n_y, n_z)$, the region-level 3D descriptor-based feature vector as $\textbf{f}_{des} \in \mathbb{R}^{F \times 1}$. $F$ is the dimension of the feature vector and it depends on the category of 3D descriptors mentioned above. The cosine angle between the normals of two adjacent regions $R_i$ and $R_j$ is denoted as $A_n$, and cosine angle between two 3D descriptors-based feature vectors of the two adjacent points/regions is denoted as $A_{des}$, the \textit{Affinity} $A(R_i, R_j)$ is given as:

\begin{scriptsize}
\begin{equation}
\centering
    A(R_i, R_j)=\sqrt{\lambda_nA_n^2+\lambda_{des}A_{des}^2},
\end{equation}
\end{scriptsize}

where the parameters $\lambda_n$ and $\lambda_{des} \in (0, 1]$, and we set $\lambda_n=\lambda_{des}=1$ in all our experiments. In the process of region merging, regions with the similarity larger than $cos\theta_{ts}$ ($cos\theta_{ts} \in (0, 1)$) are merged iteratively. The threshold $\theta_{th}$ is very important, which determines whether or not a supervoxel should be merged in the next cluster-level region merging. In all our experiments, we have set the $\theta_{th}$ to $60^{\circ}$. The detailed process of region merging is given in Algorithm \ref{alg_clustering}, and is summarized as follows:
\begin{algorithm}[t!]
\footnotesize
     \caption{The Learning-based Region Merging based on 3D descriptors} 
     \label{alg_clustering}
      \KwIn{The \textbf{input} region set $\textbf{R}_i=\{r_i\}, i=1, 2, ..., N_{i}$. The region set contains regions with labeled points and regions with the top 2\textperthousand\ minimum curvature.} 
      \KwOut{The \textbf{output} pseudo label matrix $\textbf{\textit{L}}_{region}$ for different regions.}
  \label{alg_1}
  Initialize  $R_{seed}=r_{\{seed, i\}}, i=1, 2, ..., N^{seed}$\; Initialize the pseudo label matrix $\textbf{\textit{L}}_{region}$ as a zero matrix.
  \While{not converged}
{ Select $K$ nearest neighbour regions $r_j$ around the seed region $r_{seed}$ for comparisons based on fast Octree-based $K$ Nearest Neighbor Search\;
        
        \For{the seed regions $r_j$ selected} 
        {
            \If{\textbf{Condition 1}}
            {Assign $r_j$ the same class label as $r_{seed}$\;
            
                \If{\textbf{Condition 2}}
                  {Regard the region $r_j$ as new seed regions\;}
            }
             \Else{Assign $r_j$ with a new class label. Regard  $r_j$ as new seed regions;} 
             $j \leftarrow j+1$\;
             Update the class pseudo label matrix $\textit{\textbf{L}}_{region}$.}
        } 
\Return The class pseudo label matrix $\textit{\textbf{L}}_{region}$ of $\textbf{R}_{i}$ with pseudo label of different regions.
\end{algorithm}

Firstly, regions are ranked according to curvatures. The regions with labeled point and regions that have top 2\textperthousand\ minimum curvature among all points are regarded as seed regions. The fast Octree-based K-nearest neighbor (KNN) search of seed regions is adopted in each iteration for acceleration. Improved based on it, we propose the following three criteria for a faster KNN region query. \textbf{1}. If an octant is not overlapped with the query ball, we skip it. \textbf{2}. If the query ball is inside an octant, we stop searching. \textbf{3}. If the query ball contains the octant, we just compare the query with all regions, so going into children of that octant is not required. We greatly improve the query speed by 18.2 times for a scene of about $3 \times 10^6$ point for example, and it also substantially speeds up the region-level average normal and curvature calculations. As shown in Algorithm \ref{alg_clustering}, we summarize the detailed algorithms for PFH-based over-segmentation as the following five steps:
\begin{enumerate}
    \item Select regions with labeled point the regions that have minimum curvatures as the initial seed regions.
    \item Utilizing fast KNN search, we obtain the neighbouring regions of the initial seed regions {$r_{seed}$}, and calculate the \textit{Affinity} $A(r_{seed}, r_j)$ of the query center region $r_{seed}$ and the neighbouring regions $r_j$. 
    \item If the local feature affinity is large enough (\textbf{Condition 1}), we assign the neighbouring region $r_j$ with the same label as $r_{seed}$. 
    \item  If \textbf{Condition 1} is satisfied, denote the curvature of $r_j$ and $r_{seed}$ as $r_j$ and $r_{seed}$, and denote the the difference of $r_j$ and $r_{seed}$ as $ \Delta r= \|r_j-r_{seed}\|$. If $ \Delta r \leq \zeta$ (\textbf{Condition 2}), we set $r_j$ as the seed region. That means we set the regions $r_j$ as seed regions, only if the difference in curvature is small enough. Otherwise, we only assign $r_j$ the same class label as $r_{seed}$.
    \item  If \textbf{Condition 1} is not satisfied, we assign the neighbouring region $r_j$ with a different label from $r_{seed}$, and also assign $r_j$ as the seed region.
\end{enumerate}
Finally, the regions with large similarities are merged with the training of the network. The  algorithm is summarized in Algorithm \ref{alg_clustering}. The loop will terminate if any of the following convergence conditions is satisfied, which are proposed as:
 \begin{enumerate}
     \item All regions have been assigned with labels;
     \item There are no seed regions that can be added;
     \item Regions will not expand between two successive steps.
 \end{enumerate}

These three conditions are designed to assign as many regions with pseudo labels as possible, which also facilitates the following self-training and region-based neural network processing.

\begin{figure*}[t!]
\centering
\includegraphics[scale=0.143]{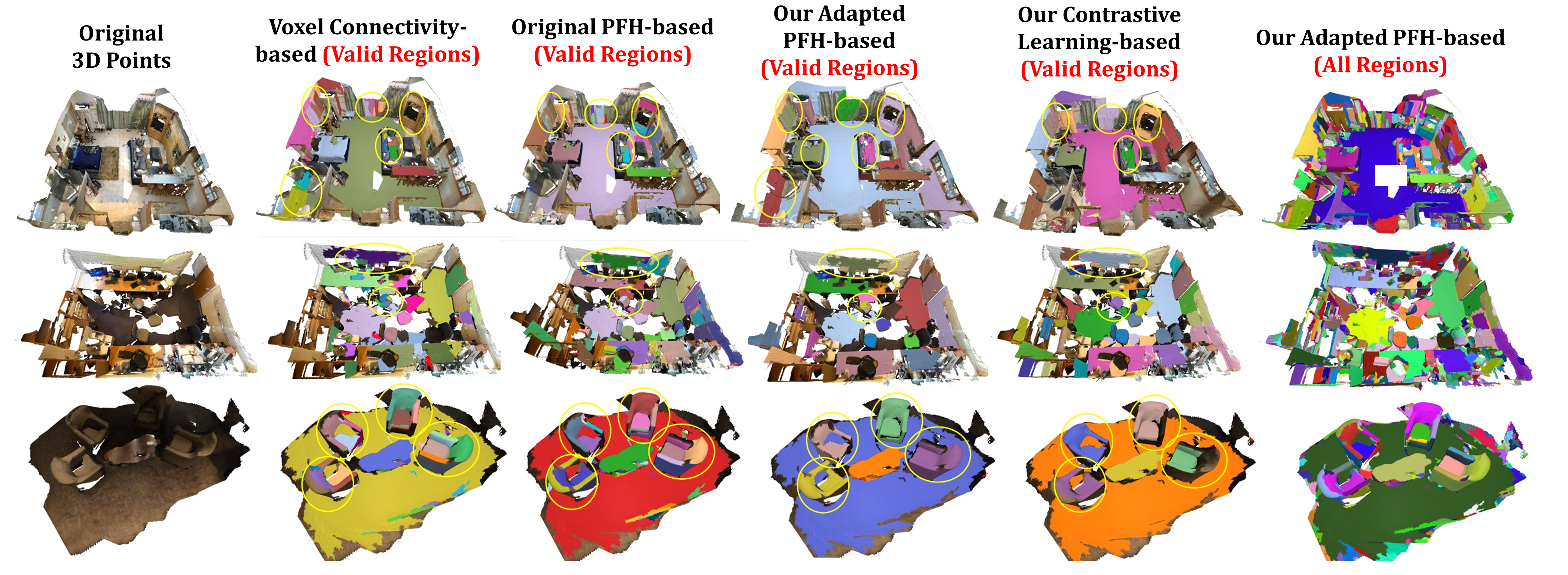}
\caption{Qualitative experimental results just after the cluster-level region merging of various approaches for dense indoor point clouds of ScanNet semantic segmentation. Note that merely valid regions with highly confident cluster-level predictions are used as pseudo labels during self-training. Compared with the original PFH-based approach and the voxel connectivity-based approach, our adapted PFH-based and contrastive learning-based approach can both provide more homogeneous and consistent region merging results. Also, it can be demonstrated that the region merging results of the original PFH-based 3D descriptor is not that good because it takes the point distance into consideration. The final semantic segmentation mean Intersection over Union (mIoU) and instance segmentation Average Precision (AP@50\%) are also comprehensively reported in Table \ref{table_inseg}, respectively. It can be demonstrated that both our adapted PFH-based and contrastive learning-based approaches show superior performance.}
\label{fig_oseg_scannet}
\vspace{-0.002cm}
\end{figure*}

As shown in Figure \ref{fig_backbone}, the output of backbone network gives the prediction of semantic segmentation with $\textbf{P}_{out} \in \mathbb{R}^{N_i \times C_{Seg}}$, where $C_{Seg}$ denotes the number of semantic categories and $N_i$ is the number of input points. For the limited annotation case, which means there are only a few annotated points (i.e. 0.2\%) in a scene with approximately $2 \times 10^6$ points, we propose the region-level similarity prediction strategy by simply adding a max pooling operation after our backbone network to offer the region-level predictions as shown in Figure \ref{fig_backbone}. To be more specific, we have adopted $1 \times 1$ convolution at the last layer of backbone network to obtain a feature $\textbf{P}_{out} \in \mathbb{R}^{N_{i} \times C_{Seg}}$, then we adopt max-pooling for each region to obtain the region-level prediction $\textbf{P}_{clu} \in \mathbb{R}^{N_{clu} \times C_{Seg}}$, where $N_{clu}$ is the initial number of regions obtained from over-segmentation. For instance segmentation, we obtain the region-level prediction $\textbf{P}_{Ins} \in \mathbb{R}^{N_{clu} \times C_{Ins}}$ for each instance based on point clustering method proposed in PointGroup, where the $C_{Ins}$ is the number of instance categories. Note that normalized scores in prediction of the merged clusters is added in each training iteration based on the similarity scores in both geometry and semantics among them. The calculation of the similarity scores in geometry/semantics is detailed in the following paragraph.

\noindent
\textbf{Learning Based Region Merging} As shown in Figure \ref{Fig_Region_Expansion}, the initial clusters obtained from over-segmention suffer from excessive dividing or inaccurate partitioning. It is desired that a region merging submodule should be proposed to merge or divide clusters in a learnable way. In our design, the predicted semantic/instance of the learnable network and the 3D traditional or learnt 3D descriptor of clusters jointly decide a similarity score, indicating whether neighbouring clusters should be merged. In 2D computer vision, the idea of using geometric feature to serve as local descriptors for object detection has been proposed. In our work, we have incorporated the 3D local descriptor into consideration to evaluate the local geometric similarities. More specifically, the similarity score between the $j_{th}$ neighbouring point cluster with prediction of $\textbf{P}_{clu,j}$ and the $i_{th}$ point cluster with prediction of $\textbf{P}_{clu, i}$ is calculated as: 

\begin{figure}[t!]
\centering
\includegraphics[scale=0.1580]{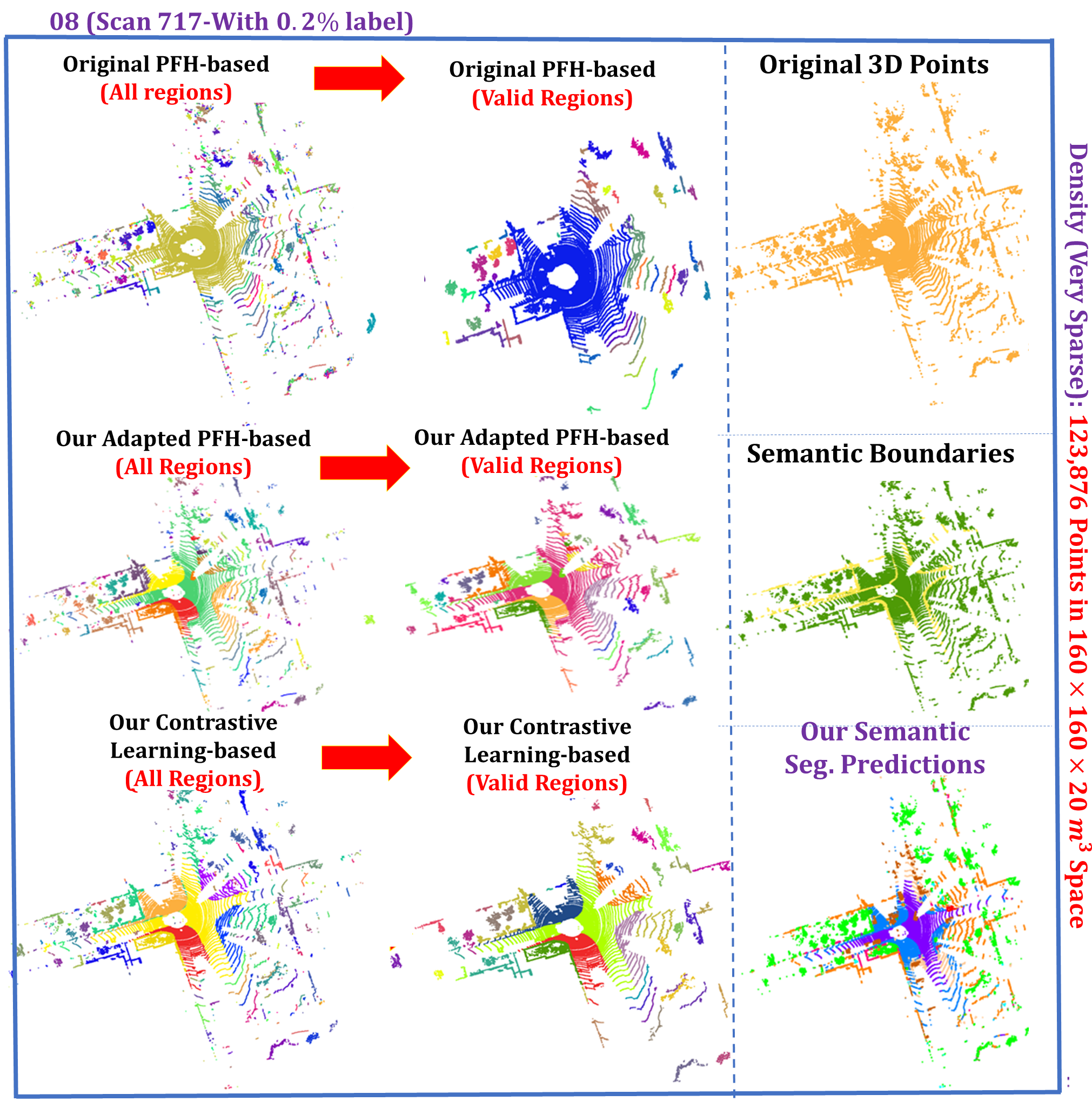}
\caption{Qualitative experimental results just after the cluster-level region merging of various approaches for the sparse outdoor LiDAR point clouds of SemanticKITTI semantic segmentation. We have shown the oversegmentation results of our adapted PFH-based approach and our proposed contrastive learning-based approach compared with the original PFH-based approach. Note that merely valid regions with highly confident cluster-level predictions are used as pseudo labels during self-training. Compared with the original PFH-based approach, both our adapted PFH-based and contrastive learning-based 3D local description approaches can provide more homogeneous and consistent region merging results. The final instance segmentation performance of SemanticKITTI is reported in Table \ref{table_inseg}. It can be demonstrated that both our adapted PFH-based and contrastive learning-based 3D local description approaches show superior performances in the tasks of instance segmentation as shown in Table \ref{table_inseg}.}
\label{fig_oseg_semantickitti}
\vspace{-0.32cm}
\end{figure}

\begin{scriptsize}

\begin{equation}
    S_{i, j}(\textbf{P}_\textit{\textbf{clu,i}}, \textbf{P}_\textit{\textbf{clu,j}})= y_1 M_{\mbox{\textit{\tiny{{color,i,j}}}}} +  y_2 M_{\mbox{\textit{\tiny{{scale,i,j}}}}}+ y_3 M_{\mbox{\textit{\tiny{{des,i,j}}}}}+ y_4 M_{\mbox{\textit{\tiny{{seg,i,j}}}}},
\end{equation}

\end{scriptsize}

where $M_{\mbox{\textit{\tiny{{color,i,j}}}}}, M_{\mbox{\textit{\tiny{{scale,i,j}}}}},  
M_{\mbox{\textit{\tiny{{des,i,j}}}}}, M_{\mbox{\textit{\tiny{{seg,i,j}}}}} \in [0, 1]$. The $M_{\mbox{\textit{\tiny{{color,i,j}}}}}$, $M_{\mbox{\textit{\tiny{{scale,i,j}}}}}$, and $M_{\mbox{\textit{\tiny{{des,i,j}}}}}$ are the scores that are the normalized average intrinsic color, dimension, and 3D local descriptor-based similarities between the point cluster $i$ and the point cluster $j$, respectively. The average 3D local descriptor-based similarity $M_{\mbox{\textit{\tiny{{des,i,j}}}}}$ is essentially the similarity of the region-level feature vector between the region $i$ and region $j$. The $M_{\mbox{\textit{\tiny{{des,i,j}}}}}$ is calculate in the same way as the PFH-based affinity $A(R_i, R_j)$ calculation in Subsection~\ref{sub_sec_merge}. While the semantic similarity $M_{\mbox{\textit{\tiny{{seg,i,j}}}}}$ between the $i_{th}$ and $j_{th}$ region is evaluated based on the output similarity of the two clusters:

\begin{scriptsize}
\begin{equation}
M_{seg}= exp\{ -\lambda \Vert\textbf{p}_{clu, i}- \textbf{p}_{clu, j}\Vert^2\},
\end{equation}
\end{scriptsize}
\\
where $\textbf{p}_{clu, i}$ and $\textbf{p}_{clu, j}$ are the corresponding predictions in $\textbf{P}_{clu}$ for neighbouring cluster $i$ and cluster $j$. We have designed \textit{the weight balancing strategy} to avoid the noisy and low-quality pseudo labels at the beginning of training. The balancing weights $y_1, y_2, y_3 \in [0, 1]$ are set to values declining from a high value to a small value, i.e. $y_1=y_2=y_3=1-\frac{m_i}{{{N_{Total}}}}$, while $y_4 \in \{0, 1\}$ is set to the value $\frac{m_i}{N_{Total}}$. Where $m_i$ represents the current iteration in training, and $N_{Total}$ is total number of training iterations. This design means we firstly trust more on similarity of the local geometric properties and gradually trust more on the updated semantics relations in cluster-level predictions. We replace the \textbf{\textit{Condition 1}} in Algorithm \ref{alg_clustering} with the \textbf{\textit{Condition 3}}: $S_{i, j} \geq 1.25$ \& $\textbf{p}_{clu, i} \geq \gamma$ \& $\textbf{p}_{clu, j} \geq \gamma$. And we substitute \textbf{\textit{Condition 2}} with  \textbf{\textit{Condition 4}}: $S_{i, j} \geq 1.5$, respectively. $\gamma$ is a confidence threshold to ensure that merely highly confident network predictions can be utilized for the network optimizations. Also, the K nearest neighbour in Algorithm \ref{alg_clustering} is conducted at cluster level. This design, combined with \textit{cluster-level similarity prediction strategy}, utilizes weak labels as the guidance to increase the quality of generated pseudo label in training iteratively for both semantic and instance segmentation tasks. And the updated pseudo labels are shown in the third column of Figure \ref{Fig_Region_Expansion} for ScanNet instance segmentation. From our further experiments in ablation studies, the traditional or learnt 3D descriptors play a great importance in capturing the local geometric properties, thus enhancing the 3D scene understanding performance. \\
\textbf{Data Augmentation Submodule}
This submodule is inspired by a simple intuition that the network prediction should be consistent under diverse transformations including flipping, rotation, and even down-sampling. The details of data augmentation is provided in the Appendix. Based on the backbone network, for the original point clouds input $\textbf{P}_{in}$ and augmented point clouds input $\textbf{P}^{aug}_{in}$, we firstly obtain the final network semantic/instance predictions $\textbf{P}_{out}, \;\textbf{P}^{aug}_{out} \in \mathbb{R}^{N_{i} \times C_{Seg}}$ respectively. For the task of instance segmentation, the outputs are $\textbf{P}_{Ins}, \;\textbf{P}^{aug}_{Ins} \in \mathbb{R}^{N_{i} \times C_{Ins}}$ respectively. The \textit{KL} divergence is universally adopted to evaluate the difference between two probabilistic distributions. In our work, we utilize the \textit{JS} divergence instead because of its symmetry property, which means it remains constant when two distributions are very distant to each other. Also, the distributional loss for regression such as the JS divergence is easier to optimize with improved gradient and reduces the overfitting problem compared with the mean square error (MSE)-based loss. It is also demonstrated by experiments that the JS divergence outperforms traditional mean square error-based loss in downstream scene understanding tasks. The final \textit{JS} Divergence Loss for semantic segmentation is formulated as:

\begin{scriptsize}
\begin{equation}
\begin{aligned}
    L^{js}_{Aug} =& -\frac{1}{N_{com}}\sum_{i=1}^{N_{com}} \textbf{Div}_{\small{JS}} (\sigma(\textbf{P}_{out})\|\sigma(\textbf{P}^{aug}_{out})),
\end{aligned}
\end{equation}
\end{scriptsize}

where $N_{com}$ is the number of random sampled common intersectional points between $\textbf{P}_{out}$ and $\textbf{P}^{aug}_{out}$. And the same goes for instance segmentation. We keep the number of random sampled points to 1000 in all our experiments for the efficiency consideration. And $\sigma$ is the Softmax function with normalization to produce probabilistic scores for each class. After applying the data augmentation constraints, we aim at ensuring that the distribution of the probabilistic scores of segmentation will remain consistent between the  $\textbf{P}_{out}$ and  $\textbf{P}^{aug}_{out}$ after various data transformations. To be more specific, the transformation invariance can be achieved.
\\
\noindent \textbf{Pseudo Segmentation Submodule}
Finally, the network is also guided by the generated pseudo label in both semantic and instance segmentation tasks, which can be formulated as:
$ L^{sem}_{\mbox{\textit{\tiny{{WSL}}}}}=\frac{1}{N_{i}}\sum_{i=1}^{N_{i}}\textit{\textbf{CE}}(\textbf{P}_{out}, \textbf{P}_{gt}) \mathds{1}(\textbf{p}_{i})$. Where $\textbf{P}_{out}$ is the segmentation output prediction, and $\textbf{P}_{gt}$ is the ground truth supervision provided by the generated pseudo labels in each training iteration. $\mathds{1}(\textbf{p}_{i}) \in \{0, 1\}$ indicates whether the point has been given a pseudo label in the current training iteration. The final optimization takes losses from all above-mentioned submodules into account, formulated as $L_{Seg}=L^{js}_{Aug}+L^{sem}_{\mbox{\textit{\tiny{WSL}}}}$. The self-training is used for network learning with pseudo labels, and the network is optimized in an end-to-end manner for semantic/instance segmentation. 

\vspace{-3mm}


\subsection{Proposed Region Merging Strategies for Detection}
The object detection network is designed based on widely adopted VoteNet. Based on VoteNet, we propose \textbf{Dice} loss to guarantee tighter aggregations of points within the same cluster, and strict geometric separations of points in diverse clusters. Note that for the object detection, our method operates in an unsupervised manner for doing instance segmentation, followed by our regression submodule to realize object detection. Note that the same as the semantic/instance segmentation branch, the pseudo segmentation submodule is utilized to perform instance segmentation, and the data augmentation submodule is utilized to ensure the transformation invariance. Different from segmentation branch, $\textbf{P}_{out} \in N_i \times (C_{det}+1)$, where $(C_{det}+1)$ is the number of object classes plus one for backgrounds in detection. Similarly, in order to obtain the object-level prediction, We apply max-pooling to $\textbf{P}_{out}$ to obtain $\textbf{P}_{det} \in N_{\mbox{\textit{\tiny{{R}}}}} \times (C_{det}+1)$, where $N_{\mbox{\textit{\tiny{{R}}}}}$ is the number of objects which are given true labels rather than pseudo labels. Then we add $1\times 1$ convolution and max pooling after $\textbf{P}_{det}$ to produce $\textbf{P}_{cls} \in \mathbb{R}^{N_{cls}} $, and it predicts the presence of object or not with a scene:

\begin{scriptsize}
    \begin{equation}
        L_{cls}=-\frac{1}{C_{cls}}\sum_{i=1}^{C_{cls}} L_{\textit{\textbf{CE}}}(\textbf{P}_{cls}, \textbf{P}^{GT}_{cls}).
 \end{equation}
\end{scriptsize}
 
The $L_{\textit{\textbf{CE}}}$ is the cross-entropy loss, and $C_{cls}$ is the number of object classes within the scene.
In this way, object presence within a scene can serve as the supervision for similarity predictions among regions benefiting from the self-supervision provided by scene object classes. 
\\
\textbf{Regression Submodule}
Object detection can take advantage of the supervision from instance segmentation because object proposals can be directly obtained from the results of the instance segmentation. The axis tightly aligned bounding box of each instance provided by the results of instance segmentation is selected as the initialization of pseudo ground truth bounding boxes for object detection.
At the same time, the \textbf{Dice loss} \cite{li2019dice} can be utilized to evaluate intersections between the predicted regions and ground truth regions for regression purpose. Note that other submodules are the same as the semantic/instance segmentation branch. Denote the \textbf{Dice loss} as $L_{Dice}$ and the same losses as segmentation branch as $L_{Seg,2}$, the total optimization function for detection is formulated as $L_{Det}=L_{Seg,2}+L_{Dice}+L_{cls}$. Our network is optimized in an end-to-end manner on a single 1080Ti GPU for three scene understanding tasks.
\vspace{-2.8mm}

\section{Experiments}
\label{sec_experi}
\subsection{Experimental Details}
\subsubsection{Experimental Details of the Self-Training}
We adopt the self-training strategy, therefore, the training of the network requires several iterations. As is illustrated in Subsection~\ref{baseline_insseg}, our self-training is a little different from the traditional manners. More specifically, we have shown results of cluster-level region merging in Figure~\ref{Fig_Region_Expansion} and Figure~\ref{fig_oseg_scannet}. As mentioned in Subsubsection~\ref{baseline_insseg}, the self-training process is done iteratively with the following two steps: we first update the pseudo labels using our proposed 3D descriptor-based region merging strategy, and then we train our network for $\sigma$ epochs. The self-training converges after eight iterations, and more training will not bring many improvements on the final results. To be exact, the value of $\sigma$ is set to 65 ($65\times 8= 520 $ epochs) in the tasks of the semantic and instance segmentation, and $\sigma$ is set to 40 ($40\times 8= 320$ epochs) in the task of the object detection. It can be seen in Figure \ref{fig_oseg_scannet} and Figure \ref{fig_oseg_semantickitti} that we have merely given pseudo labels for those confident merged region-level predictions. The confidence threshold $\gamma$ is set to a fixed value of 0.75 in all our experiments, which is very significant to ensure the high-confidence pseudo label in region merging and the final performance of our framework. According to our ablation studies, this confidence regions based self-training strategy can have a enhancement on the final 3D scene understanding performances.
\vspace{-0.99mm}
\subsubsection{Experimental Details of the Main Network}
Unlike previous work PSD which utilizes point-based network as the backbone, in this work, we choose the voxel-based SparseConv~\cite{graham20183d} as our backbone in all our experiments for its simplicity and strong performance in both tasks of detection and segmentation. It should be noted that all our proposed network modules and loss functions are only required in training. As the training finished, the network weights are fixed in testing, and all our proposed network modules are not required.  \\


\vspace{-5mm}

For the \textbf{semantic segmentation and the instance segmentation}, the network is trained for 520 epochs on a single 1080Ti with the batch size of 8 during training and 16 during testing. The initial learning rate is $1e^{-3}$ and decays by 5 times every 60 epochs. 

For the \textbf{object detection}, we follow the training settings for the VoteNet~\cite{qi2019deep}, i.e., we use Adam optimizer with batch size of 16 and an initial learning rate of 1$\times$10$^{-3}$. The learning rate reduces by 10 times after 80 epochs and then reduces by another 10 times after 160 epochs. The total training epochs is 320.  Taking ScanNet for example, training takes 5.8 hours. We implement it in \textit{PyTorch} and optimized it with Adam optimizer. In our work, the \{20, 50, 100, 200\} labeled points represent the labeled points per scene. We adopt the random selection of label points and ensure that at least one point is selected for each class. In our experiments, we keep the same label points percentage (e.g. number of labeled points per scene) for various of compared methods, thus guaranteeing the fairness of comparisons. The random labeling scheme adopted by us is the most efficient and common data labeling scheme of point clouds in 3D WSL. In practice, there is no unified solution to obtain the certain portion of label points, as long as the labeling strategy is efficient enough. All our experiments have been done three times, and the average results are given. Although with randomly sampled label points, we still outperform all compared SOTAs methods such as the ReDAL \cite{wu2021redal} with active learning. Moreover, random selection of label points in \textit{RM3D} is far more efficient than other ones.
\vspace{-11mm}
\subsection{Quality of the Produced Pseudo Label by RM3D}
\label{sub_pseudo_label}
\begin{figure}[t!]
\centering
\includegraphics[scale=0.29588]{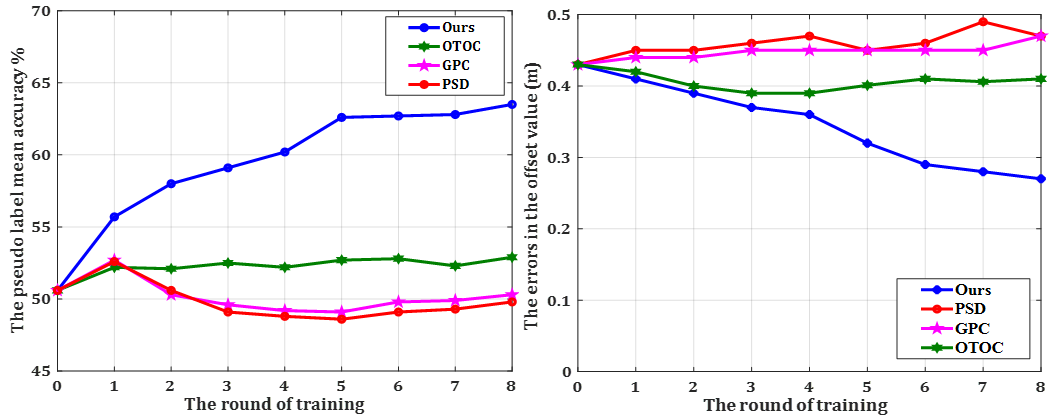}
\caption{The comparison and analysis of the pseudo label quality and its influence on the predict errors in offset in instance segmentation on the ScanNet benchmark with 0.2\% labels. Our proposed approach is both superior in pseudo label predictions and offset predictions.}
\label{fig_Label_com}
\vspace{-6.2 mm}
\end{figure}

We have also compared the quality of pseudo labels provided by diverse SOTAs approaches as shown in Figure \ref{fig_Label_com}. We unified the backbone to SparseConv \cite{graham20183d} and for a fair comparison. It is demonstrated that the key success our proposed approach lies in that can produce high-quality pseudo labels, which results in the high performance in 3D understanding tasks. 

The existing works GPC \cite{jiang2021guided}, OTOC \cite{liu2021one} focuses on contrastive learning. The network directly learns the instance discrimination capacity using the positive and negative samples, and pull the positive samples together while pushing the negative samples away in embedding space. According to our experimental results, the constraints in contrastive learning \cite{jiang2021guided, liu2021one} is too hard, because it minimizes the distance between the positive samples and maximize the distance between negative samples. While in real circumstances, the same semantic category might have some distance, while the different semantic category can be not that far away. As shown in Figure \ref{fig_Label_com}, according to our experiments, the contrastive learning-based approaches GPC and OTOC can generate low-quality pseudo label in self-training if the initial pseudo label quality is not good enough. 

While in our proposed self-training pipeline, only the similar regions with high confidence propagate labels to each other, which guarantee the quality of pseudo labels in the first few training iterations. Also, we have the following design to avoid low-quality initial labels in the self-training. In the first few training iterations, the quality of labels is low, we choose to firstly believe more in the traditional or learnt local 3D descriptor-based features $M_{\mbox{\textit{\tiny{{color,i,j}}}}}, M_{\mbox{\textit{\tiny{{scale,i,j}}}}},$ and $M_{\mbox{\textit{\tiny{{des,i,j}}}}}$ to do the region merging. And the quality of the pseudo label becomes higher with the increasing of training iterations, then we trust more on the semantic similarity $M_{\mbox{\textit{\tiny{{seg,i,j}}}}}$ to update the pseudo labels. According to our ablation studies, this strategy has successfully avoided the low-quality initial labels and improved the final performances.

\begin{table*}[htbp]
\begin{center}
\resizebox{\linewidth}{!}{\begin{tabular}{l|lll|lll|lll}
\toprule [1.5pt]
\multirow{2}{*}{Case No.} & \multicolumn{3}{c|}{S3DIS Area 5} & \multicolumn{3}{c|}{ScanNet}  & \multicolumn{3}{c}{v-KITTI 6 fold} \cr &  Recall& Precision & F1 Score & Recall& Precision & F1 Score & Recall & Precision & F1 Score\\
\midrule[1pt]                   
Visual-Similarity \cite{chen2003visual}& 62.6/51.8  & 23.1/17.5 & 33.7/26.1 &46.5/40.9&10.2/8.1&16.7/13.5 & 63.3/59.6 & 15.6/11.2 & 25.0/18.9 \\

HKS \cite{sun2009concise}& 65.9/63.8 & 27.9/26.5 & 39.2/37.4 &48.2/46.0 &11.9/9.2 &19.1/15.3 & 65.1/63.8 & 17.4/15.6 & 27.5/25.1   \\

SIKS \cite{bronstein2010scale}& 67.3/67.1 & 28.6/28.2 & 40.1/39.7 &49.6/48.9 &12.8/12.3 &20.3/19.7 & 66.9/66.5 & 18.6/18.1 & 29.1/28.5 \\

F-PFH \cite{rusu2009fast}& 68.8/68.5 & 29.3/28.6 & 41.2/40.4 &50.6/49.8 &15.8/16.7 &24.1/25.0 & 68.1/67.8 & 19.3/18.8 & 30.0/29.5   \\

WKS \cite{aubry2011wave}& 69.4/68.6 & 29.9/29.1 & 41.8/40.9 & 51.3/50.6 & 16.5/15.9 & 25.0/24.2 & 68.4/67.8 & 20.2/19.8 & 31.2/30.6   \\

SHOT \cite{salti2014shot} & 70.2/69.7 & 30.6/30.0 & 42.6/41.9 &52.1/52.0 & 17.4/16.8 & 26.1/25.4 & 69.3/68.9 & 21.0/20.6 & 32.2/31.7 \\

Vox-Connect \cite{papon2013voxel}& 70.7/69.2 & 31.2/28.7 & 43.3/40.6 &52.6/52.3 & 18.1/17.9 & 26.9/26.7 & 70.0/69.6 & 21.8/21.5 & 33.2/32.9
\\
Superpoint-Graph \cite{landrieu2018large}& 71.1/70.2 & 32.3/31.0 & 44.4/43.0 &53.1/52.5 &18.2/17.7 & 27.1/26.5 & 70.2/69.3 & 23.1/22.7 & 34.8/34.2\\

Original PFH \cite{rusu2008aligning}& 71.2/70.1 & 32.2/31.8 & 44.3/43.8 & 52.8/52.3 & 18.3/18.0 &27.2/26.8 & 70.5/69.7 & 22.3/22.0 & 33.9/33.4\\

Adapted PFH \textcolor{red}{(Our)} & 72.1/72.0  & 32.4/32.2 & 44.7/44.5 &53.8/53.5 &18.8/18.4 & 27.9/27.4 & 71.1/70.8 & 23.2/23.0 & 35.0/34.7\\

Point-SIFT \cite{jiang2018pointsift}& 71.2/69.2 & 32.6/31.3 & 44.8/43.1 &53.5/52.2 &19.0/17.3  &28.0/26.0 & 70.9/69.6 & 23.3/21.6 & 35.1/33.0\\

S-SPG \cite{landrieu2019point} & 72.8/72.3  & 34.2/33.9 & 46.5/46.2 &55.0/54.2 & 19.8/19.3 & 29.1/28.5 & 72.3/72.0 & 25.0/24.8 & 37.2/36.9\\

Contrastive Learning \textcolor{red}{(Our)} & 74.5/73.9 & 37.3/36.7 & 49.7/49.1 &54.1/53.9 & 19.2/18.8 & 28.3/27.9 & 71.2/70.9 & 24.1/23.7 & 36.0/35.5\\

Predator \cite{huang2021predator} & 75.2/72.9  & 37.1/35.8 & 49.7/48.0 & 55.3/53.8 & 20.2/18.7 & 29.6/27.8 & 72.2/71.8 & 25.0/24.1 & 37.1/36.1\\
\bottomrule [1.5pt]
\end{tabular}}
\caption{The performance of various 3D descriptors in oversegmentation tasks on S3DIS, ScanNet~\cite{dai2017scannet}, and v-KITTI \cite{gaidon2016virtual} for the original (Left) and rotated (Right) scenes, respectively. Left of '/': Original. Right of '/': Rotated.  We adopt random rotation of $0 ^{\circ}$ to $180 ^{\circ}$, and then evaluate the quality of over-segmentation in terms of rotational robustness. The backbone utilized for comparisons is unified as SparseConv \cite{graham20183d}. We test the performance of oversegmentation in the 0.2\% annotation percentage setting.}
\label{table_des_no_ro}
\vspace{-3mm}
\end{center}
\end{table*}

\begin{table*}[t!]
\begin{center}
\resizebox{\linewidth}{!}{\begin{tabular}{l|ccccc|ccccc}
\toprule [1.5pt]
\multirow{2}{*}{Case No.} & \multicolumn{5}{c|}{ScanNet AP 50\%} & \multicolumn{5}{c}{S3DIS AP 50\%}   \cr & 0.2\% & 0.4\% & 0.6\% & 0.8\% & 1.0\% &  0.2\% & 0.4\% & 0.6\% & 0.8\% & 1.0\%\\
\midrule[1pt]                     
Visual-Similarity \cite{chen2003visual} & 46.5/44.3  & 52.6/47.8 & 53.2/48.6 & 54.3/49.3 & 55.3/50.1 & 45.9/40.6 & 46.5/42.1 & 46.8/42.5 & 47.5/42.9 & 49.2/44.8 \\

HKS \cite{sun2009concise} & 51.1/49.8 & 53.4/51.9 & 54.3/52.5 & 55.1/53.2 & 55.9/53.9 & 50.2/45.8 & 51.3/48.9 & 52.2/50.5 & 53.1/51.2 & 53.9/52.3 \\

SIKS \cite{bronstein2010scale}  & 51.9/51.5 & 52.9/52.5 & 53.4/52.9 & 54.2/53.8 & 54.9/54.7 & 52.7/52.1 & 53.2/52.7 & 53.9/53.2 & 56.0/55.3 & 56.8/56.5\\

F-PFH \cite{rusu2009fast} & 52.6/52.0 & 53.2/52.7 & 54.2/53.8 & 55.2/54.8 & 56.0/55.7 & 52.9/52.5 & 55.3/54.9 & 55.9/55.3 & 56.8/56.5  & 57.1/56.8\\

WKS \cite{aubry2011wave} & 53.2/52.7 & 53.5/53.2 & 54.2/53.7 & 54.9/54.6 & 55.5/55.3 & 54.2/53.8 & 55.2/54.9 & 56.9/56.5 & 56.6/56.3 & 57.4/56.9 \\

Vox-Connect \cite{papon2013voxel} & 54.0/53.7 & 54.9/54.5 & 55.7/55.3 & 56.2/55.6 & 57.0/56.5 & 53.7/53.1 & 54.2/53.6 & 54.8/54.3 & 56.6/56.1 & 57.6/56.9 \\

SHOT \cite{salti2014shot} & 54.5/53.6 & 55.2/54.7 & 56.1/55.6&57.2/56.7 &57.9/57.4 & 54.4/53.9 & 54.9/54.5 & 55.8/55.5 & 56.7/56.2 & 57.8/57.3
\\

Superpoint-Graph \cite{landrieu2018large} & 54.7/53.0 & 55.3/53.5 & 55.6/53.9 & 55.9/54.1 & 57.3/55.9 & 54.3/52.1 & 54.9/53.0 & 55.7/54.6 & 56.8/55.3 & 58.2/56.9 \\

Original PFH \cite{rusu2008aligning} & 55.1/54.7  & 55.9/54.9 & 56.4/55.6 & 57.2/56.6 & 58.1/57.8 & 54.7/53.9 & 55.1/54.8 & 55.9/55.5 & 57.3/57.0 & 58.3/58.5 \\

Adapted PFH \textcolor{red}{(Our)} & 55.6/55.1 & 56.1/55.6 & 56.7/56.6 & 57.9/57.6 & 58.6/58.2 & 54.9/54.3 & 55.3/55.2 & 56.0/55.8 & 57.9/57.5 & 58.8/58.6\\

Point-SIFT \cite{jiang2018pointsift}  & 55.9/55.3 & 56.4/56.1 & 56.8/56.6 & 57.4/57.0 & 58.0/57.5 & 54.9/53.5 & 55.1/54.7 & 56.1/55.7 & 56.9/56.7 & 58.6/58.4\\

S-SPG \cite{landrieu2019point} & 56.9/56.5 & 57.8/57.3 & 58.3/58.2 & 58.5/58.7 & 59.1/59.3 & 55.9/55.5 & 56.3/56.1 & 57.3/57.1 & 57.9/57.5 & 59.9/59.2\\
 
Contrastive Learning \textcolor{red}{(Our)} &57.6/57.2 & 57.9/57.7 & 58.5/58.4 & 58.9/58.7 & 60.5/59.8 & 56.3/55.5 & 56.7/55.9 & 57.2/56.5 &57.9/57.0 & 60.8/59.6\\

Predator \cite{huang2021predator}  & 57.9/56.6 & 58.3/57.2 & 59.2/57.8 & 59.8/58.3 & 61.5/58.6 & 57.0/55.3 & 57.3/55.6 & 57.8/55.9 & 58.3/56.3 & 61.3/59.2 \\
\bottomrule  [1.5pt]   
\end{tabular}}
\caption{The performance of various 3D descriptors in instance segmentation tasks on S3DIS, ScanNet for the original and rotated scenes, respectively. Left of '/': Original. Right of '/': Rotated. The tested annotation percentages are 0.2\%, 0.4\%, 0.6\%, 0.8\%, and 1.0\% for the task of instance segmentation. The labeled points are randomly selected. We adopt random rotation of $0 ^{\circ}$ to $180 ^{\circ}$, and then evaluate the quality of instance segmentation in terms of rotational robustness. The backbone utilized for comparisons is unified as the SparseConv \cite{graham20183d}.}
\label{table_inseg}
\vspace{-5mm}
\end{center}
\end{table*}

Finally, we have shown the quality of the pseudo labels provided by our proposed region merging based approach compared with the SOTA methods OTOC \cite{liu2021one}, GPC \cite{jiang2021guided}, and PSD. The GPC works for the limited reconstruction case, we have extended the method to make it suitable for the limited annotation case. As shown in Figure \ref{fig_Label_com}, our proposed method has outperformed the state-of-the-art approaches OTOC \cite{liu2021one}, GPC \cite{jiang2021guided}, and PSD by a large margin for both the pseudo label generation and the offset predictions. Therefore, our proposed approach can provide more high-quality and noise-free pseudo labels for the downstream 3D scene understanding tasks. The high-quality and noise-free pseudo labels are certainly beneficial for the scene understanding downstream tasks for the fact that more correctly labeled points are used for the self-training. It demonstrates the superior effectiveness of our proposed region merging design.
\subsection{A Comprehensive Comparison on the Results of Traditional and our Proposed Learnt 3D Descriptors}

A main component of our proposed network framework is the 3D descriptor which capture the geometries of the local structure. In this Subsection, we have given a very comprehensive comparisons of the performance of different local descriptors not only on the task of over-segmentation, but also on the task of final semantic segmentation, instance segmentation, and object detection. First of all, we compare different methods on the task of over-segmentation. To test the rotational robustness of various 3D local descriptors, we have done detailed experiments for the original scene and the rotated scene respectively.  We adopt random rotation of $0 ^{\circ}$ to $180 ^{\circ}$, and then evaluate the quality of over-segmentation, semantic segmentation, and instance segmentation in terms of rotational robustness. The experimental results of instance segmentation are shown in Table \ref{table_inseg}. The left of '/' shows the performances for the original scene, and the right of '/' shows the performances for the rotated scene. It can be seen that the above illustrated 3D descriptors can be integrated seamlessly to our proposed weakly supervised learning framework to fulfill the tasks of 3D scene understanding. According to our experimental results, the visual similarity based 3D descriptor \cite{chen2003visual} has poor performance in the tasks of oversegmentation, semantic segmentation, and instance segmentation. It can be explained by the fact that the visual similarities of the 3D structures from diverse angles can be not that similar, and it results in poor local geometrical description capacity and rotation robustness. Also, the signature based approaches including HKS \cite{sun2009concise} and WKS \cite{aubry2011wave} are also not very robust to rotation because they do not explicitly consider rotational robustness in their formulations. The HKS (Heat Kernel Signiture) \cite{sun2009concise, gebal2009shape} utilizes the heat diffusion process to capture the extreme surface change. The SIKS \cite{bronstein2010scale} designs a scale-invariant heat kernel descriptor based on the diffusion scale space analyses. The performances of SIKS \cite{bronstein2010scale} for the tasks of oversegmentation, semantic segmentation and instance segmentation are marginally better than HKS \cite{sun2009concise} for its effective scale invariant heat kernel signature designs. The SIKS \cite{bronstein2010scale} is a scale-invariant version of HKS that can maintain invariance under a wide range of transformations the shape undergoes, therefore, the semantic segmentation and instance segmentation performances of SIKS \cite{bronstein2010scale} are maintained even if randomly rotated. WKS \cite{aubry2011wave} embeds and separates information from diverse Laplacian eigen frequencies by varying the energy of the quantum mechanical particle.  The WKS \cite{aubry2011wave} is invariant to isometries and very robust to small non-isometric deformations compared with HKS. Also, for the fact that WKS \cite{aubry2011wave} permits access even to very high frequency information, it results in marginally better local description capacity and provides more accurate local matching and registration than the HKS. Our experimental results also demonstrates the effectiveness of WKS compared with HKS. The results of oversegmentation and 3D scene segmentation of WKS is comparable to SIKS as shown in Table \ref{table_inseg}. 

\begin{table*}[t]
\setlength{\abovecaptionskip}{-0cm}
\setlength{\belowcaptionskip}{-0cm}
\tiny
\begin{center}
\resizebox{\linewidth}{!}{\begin{tabular}{c|c|ccc|ccc|cc|ccc}
\toprule [1.5pt]
\multirow{2}{*}{Settings} &\multirow{2}{*}{Method} & \multicolumn{3}{c|}{ScanNet Semantic Seg.\%} & \multicolumn{3}{c|}{ScanNet Instance Seg.\%} & \multicolumn{2}{c|}{S3DIS Semantic Seg.\%}& \multicolumn{3}{c}{S3DIS Instance Seg.\%} \cr & &mIoU\% & Bathtub & bed & AP@50\% & bed & bookshelf &mIoU\%& wall& mPrec\% &mRec\% & AP@50\%
\\
\midrule[1pt]   
\multirow{8}{*}{0.2\%}&WeakLabel-3DNet \textcolor{red}{(Our)} & \textcolor{red}{\textbf{65.2}}&79.6&77.5&\textcolor{red}{\textbf{55.1}}&69.3&48.7&\textcolor{red}{\textbf{66.2}}&69.5&65.8&48.3 &\textcolor{red}{\textbf{54.3}}\\
&ReDAL\cite{wu2021redal}&63.7 & 77.1 &75.1 & 51.6&64.3&46.1&63.5&62.8&62.5&46.9&52.2\\

&PSD&61.5 & 74.9 &73.7 & 50.1&58.5&42.3&58.8&57.6&60.3&44.5&49.6\\

&Xu. et al. \cite{liu2021one}&57.6&71.7 &71.3 & 47.9 &56.9 &41.6&55.9&53.8&57.5&42.3&45.7\\

&Viewpoint-Bottleneck \cite{luo2021pointly}&55.3&68.1&67.7&46.6&56.1&41.3&54.4&51.5&55.3&40.9&42.2\\

&ContrastiveSceneContext (CSC) \cite{hou2021exploring}&54.2&67.5&63.9&43.4&53.5&40.8&51.3& 49.8&53.6&38.5& 40.4\\

&PointContrast \cite{xie2020pointcontrast}&52.3&65.9&61.5&41.9&51.8&38.7&49.5&45.5&51.6&36.6&39.0\\

&SparseConv Baseline \cite{graham20183d}&46.5 &61.9 &56.6&35.3&45.2&33.5&43.3& 39.2&46.6&32.4& 34.7\\
\midrule[1pt]
\multirow{7}{*}{One Pt. (0.03\%)}&WeakLabel-3DNet \textcolor{red}{(Our)}&\textcolor{red}{\textbf{59.6}}&67.4&65.6&\textcolor{red}{\textbf{50.3}}&63.6&43.9&\textcolor{red}{\textbf{61.5}}&63.7&60.3&43.7&\textcolor{red}{\textbf{49.6}}\\
&ReDAL\cite{wu2021redal}&57.8&65.3&63.4&46.8&57.5&39.9&58.2&58.8&57.3&41.6&47.8\\
&PSD &54.6&63.8&61.5&45.3&54.8&37.9&55.7&53.7&55.1&39.8&43.2\\
&Xu. et al. \cite{liu2021one}&51.3&62.5&60.3&44.2&51.5&36.1&52.4&52.5&54.2&38.2&40.9\\
&Viewpoint-Bottleneck \cite{luo2021pointly}&48.4&59.8&57.5&40.6&49.6&35.2&50.2&50.1&52.9&37.1&39.6\\
&ContrastiveSceneContext \cite{hou2021exploring}&47.7&58.5&56.9&37.9&47.5&34.3&47.7&46.5&50.1&36.6&38.7\\
&PointContrast \cite{xie2020pointcontrast}&45.9&54.8&51.5&35.1&46.2&32.7&44.8&44.8&48.6&35.7&37.8\\
&SparseConv Baseline \cite{graham20183d} &40.1&49.3&44.6&30.9&40.7&26.2&39.5&39.6&42.9&30.9&32.5\\

\midrule[1pt]
\multirow{7}{*}{1.0\%}&WeakLabel-3DNet \textcolor{red}{(Our)}&\textcolor{red}{\textbf{67.9}}&82.7&79.9&\textcolor{red}{\textbf{58.2}}&72.2&53.1&\textcolor{red}{\textbf{68.1}}&71.8&67.5&50.8&\textcolor{red}{\textbf{58.6}}\\
&ReDAL \cite{wu2021redal}&65.1&79.4&76.3&54.5&65.6&48.9&65.5&68.9&64.6&47.6&55.3\\
& PSD &62.7&76.9&74.8&51.9&60.9&44.8&61.7&63.1&62.7&45.9&52.3\\
&Xu. et al. \cite{liu2021one} &58.5&72.8&72.9&49.3&58.6&43.1&57.8&60.0&58.9&43.9&49.2\\
&Viewpoint-Bottleneck \cite{luo2021pointly}&57.6&71.3&68.9&47.7&57.8&42.9&55.6&58.7&57.1&44.2&46.1\\
&ContrastiveSceneContext \cite{hou2021exploring}&55.3&70.8&65.5&44.4&54.8&41.9&54.8&57.7&54.7&42.8&42.6\\

&PointContrast \cite{xie2020pointcontrast}&53.6&69.7&63.5&43.6&52.7&39.8&52.5&53.4&52.6&38.2&40.1\\

&SparseConv Baseline \cite{graham20183d}&47.6&63.8&57.9&35.8&45.7&33.7&44.5&46.6&48.4&32.9&34.3\\

\midrule[1pt]
100\%&Full-Supervised SparseConv \cite{graham20183d}&72.7&83.9&82.1&63.2&76.7&62.8&71.5&74.5&71.6&55.8&62.3\\
\bottomrule [1.5pt]
\end{tabular}}
\caption{The comparisons of the performance of our proposed method on various of \textbf{\textit{indoor benchmarks}}. For weakly supervised semantic/instance segmentation, various test settings of 0.03\%, 0.2\%, and 1.0\% labeled points are experimented on ScanNet for comparisons. The setting of "One Pt." represents merely one labeled point for each class within the whole scene instead of small blocks (e.g. $1 \times 1 \times 1$ cubic meters) of Xu et al. \cite{xu2020weakly}. The backbone utilized for comparisons is unified as SparseConv \cite{graham20183d}. All experimental results are three times on average.}
\label{tablewslresults_sem_indoor}
\end{center}
\vspace{-6mm}
\end{table*}

Furthermore, according to our experimental results shown in Table \ref{table_inseg}, our proposed simple contrastive learning-based 3D descriptor has outperformed the typical learning-based descriptors such as Point-SIFT  and S-SPG in the tasks of semantic segmentation, instance segmentation, and oversegmentation. The success of the constrative learning-based local descriptor can be ascribed to the fact that contrastive learning has the capacity of capturing very discriminative feature representations of the local 3D geometry in an unsupervised manner. Compared with other learning-based 3D descriptors such as Point-SIFT, and S-SPG mentioned in Section \ref{sec_metho}, the discrimination of the positive and negative samples in the contrastive learning can be realized in a clearer way, thus resulting in a slightly better result. We have also compared our contrastive learning-based local 3D descriptor with a recently proposed 3D descriptor Predator, which is specially designed for registering point cloud scans with a low overlap. It can be demonstrated that our proposed simple contrastive learning-based 3D descriptor can achieve a slightly inferior and sometimes comparable performance compared with the Predator, which demonstrates the effectiveness of our 3D descriptor design. Also, the rotational robustness of our proposed contrastive learning-based 3D descriptor can sometimes be even better than the Predator \cite{huang2021predator}, which can be explained by the fact that the contrastive learning has inherently guaranteed the rotational robustness because the selection of the positive and the negative samples is agnostic to viewing angles. In summary, it can be demonstrated from Table \ref{table_inseg} that our proposed framework can be integrated seamlessly with various of traditional or learnt 3D descriptors to achieve 3D scene understanding. Also, our proposed adapted PFH-based 3D descriptor and contrastive learning-based 3D descriptor have satisfactory performance for the oversegmentation and instance segmentation tasks in 3D scene understanding. It is validated by experiments that our adapted PFH-based 3D descriptor and proposed contrastive learning-based 3D descriptor also have good rotational robustness. The performance will not drop much even if the random rotation of $0 ^{\circ}$ to $180 ^{\circ}$ is applied.

Last but not the least, we have also shown the experimental results just after the learning-based region merging of various approaches. The results for indoor ScanNet are shown in Figure \ref{fig_oseg_scannet}. and the results for outdoor SemanticKITTI are illustrated in Figure \ref{fig_oseg_semantickitti}. For the SemanticKITTI, we have a very simple but effective trick to tackle low-density outdoor LiDAR points: discarding all regions containing less than $N_{ths}$ points, i.e. not using these regions in all our network modules ($N_{ths}=100$ empirically for SemanticKITTI). Thus, as shown in Figure \ref{fig_oseg_semantickitti}, isolated regions with fewer points than $N_{ths}$ are successfully discarded for reliable and robust region-level predictions. As shown in Fig. \ref{fig_oseg_semantickitti}, compared to original PFH-based 3D local descriptors, our adapted PFH-based method and contrastive learning-based method better distinguish similar semantic classes such as \textcolor[RGB]{128,10,205}{road} and \textcolor[RGB]{0,100, 255}{sidewalk}, revealing that our proposed methods can have a better local feature description capacity. It is demonstrated qualitatively that both our proposed adapted PFH-based and contrastive learning-based local description approaches can provide more homogeneous and consistent region merging results compared with the traditional PFH-based approach \cite{rusu2008aligning} and the voxel connectivity-based approach \cite{papon2013voxel}. It implies that our proposed adapted PFH-based and our proposed contrastive learning-based local description approaches can extract more discriminative feature representations of local 3D structures compared with previous approaches \cite{rusu2008aligning, papon2013voxel}. According to our experimental results in Table \ref{table_inseg}, our proposed adapted PFH-based 3D local descriptor provides better performance in oversegmentation, and instance segmentation compared with various traditional faeture descriptors including HKS \cite{sun2009concise}, SIKS \cite{bronstein2010scale}, FPFH \cite{rusu2009fast}, and WKS \cite{aubry2011wave}. Our proposed adapted PFH-based 3D feature descriptor provides comparable performance with the learning-based approaches Point-SIFT \cite{jiang2018pointsift}, and S-SPG \cite{landrieu2019point}, which demonstrates its strong 3D local geometry description capacity and its effectiveness in 3D scene understanding. Also, the performance of rotational robustness of it is superior compared with other traditional and learnt 3D descriptors.

On the other hand, our proposed contrastive learning-based 3D descriptor attains comparable performance compared with the Predator \cite{huang2021predator}. It demonstrates that our proposed contrastive learning-based approach is a good choice for describing local geometry. 
In summary, it can be demonstrated in Table \ref{table_inseg} that our proposed learning-based region-merging approach can be integrated seamlessly with various traditional or learnt 3D descriptors to achieve effective 3D scene understanding. In the following experiments, we choose to use our proposed adapted PFH-based 3D local descriptor to conduct region merging for its high efficiency and for the fact that it does not need additional training data and can operate in an unsupervised manner. 
\subsection{Results of WSL for 3D Semantic/Instance Segmentation}
\vspace{-18mm}
\subsubsection{Semantic Segmentation}


We have tested our framework extensively on various large-scale point clouds understanding benchmarks including indoor S3DIS, and ScanNet \cite{Hou_2019_CVPR} and outdoor SemanticKITTI \cite{behley2019semantickitti}, and Semantic3D \cite{hackel2017semantic3d} for 3D Semantic Segmentation with limited percentage of labeled points. The 0.2\% and 1.0\% labeling settings means there are 0.2\% and 1.0\% randomly selected points that are labeled. The setting of "One Pt." in our experiments represents merely one labeled point for each class within the whole scene instead of small blocks (e.g. $1 \times 1 \times 1$ cubic meters) of Xu et al. \cite{xu2020weakly}. The results of indoor semantic segmentation are shown in Table \ref{tablewslresults_sem_indoor}, and the results for outdoor semantic segmentation are shown in Table \ref{tablewslresults_sem_outdoor}. Many previous work has explored the point cloud segmentation in the weakly supervised settings. The graph-based label propagation has both been explored by Xu et al. \cite{xu2020weakly} and OTOC \cite{liu2021one}. Xu et al. \cite{xu2020weakly} has proposed weakly supervised approach for 3D part segmentation with the settings of 1\% labeled points and the setting of labeling one point for each category. The graph-based label propagation is resembled in it based on the spatial and color smoothness constraints. The spatial and color smoothness are utilized for contrasting between diverse regions (OTOC \cite{liu2021one}) and points (Xu et al. \cite{xu2020weakly}). While our method directly utilizes both the 3D descriptors and high-level semantics for region merging rather than contrasting. Compared with them, it is demonstrated in Subsection \ref{sub_pseudo_label} that our method can generate more high-quality and noise-free pseudo labels for the segmentation tasks. It indicates that the contrastive learning-based loss in OTOC \cite{liu2021one} and spatial and color smoothness contrasting design in Xu et al. \cite{xu2020weakly} can exert too hard constraints for differentiating between positive and negative samples, and our proposed soft confidence-based region-merging can be a better choice for providing high-quality pseudo labels. The experimental results to some extent also demonstrate that our region merging strategy is better compared with exerting hard contrasting among region-level predictions. we unify the backbone of diverse approaches to SparseConv for a fair comparison.

As shown in Table \ref{tablewslresults_sem_indoor}, for the indoor ScanNet semantic segmentation, it can be demonstrated that our WeakLabel-3DNet can outperform the active learning-based approach ReDAL under the same labeling percentage. In the 0.2\% label case, our method outperforms ReDAL in semantic segmentaion by 1.5\%. It indicates we may not need the complicated active selection of the labeled points, and 3D local descriptor-based similar region merging strategy can be a better choice for considering region-level diversity. 
Compared with PSD, our method can outperform it despite sophisticated data augmentation-based consistent learning is used in it. Finally, the performance of our approach also outperforms Xu et al. \cite{xu2020weakly}, Viewpoint-Bottleneck, and pre-training-based approached such as ContrastiveSceneContext \cite{hou2021exploring} and PointContrast \cite{xie2020pointcontrast} under diverse labeling budget such as 0.2\% labels, one point label, and 1.0\% label.  On ScanNet validation set, our method has the best performance across various of objects ranging from objects spreading across a long spatial range such as the wall, and the local objects such as the bed. The experimental results prove that our method can not only capture the spatial long-range dependencies, but also the local geometric features.



As shown in Table \ref{tablewslresults_sem_outdoor}, our proposed method consistently outperforms current arts ReDAL \cite{wu2021redal} and PSD under diverse labeling percentage, which demonstrates the effectiveness of our proposed approach. In summary, it has been demonstrated by extensive experimental results in Table \ref{tablewslresults_sem_indoor} and \ref{tablewslresults_sem_outdoor} our proposed method has competitive performance for the weakly supervised semantic segmentation.
\subsubsection{Instance Segmentation}
\vspace{-2.5mm}
For the methods merely designed for semantic segmentation, we have integrated those methods with the PointGroup backbone for instance segmentation, and with Votenet backbone for object detection as our proposed approach. As shown in Table \ref{tablewslresults_sem_indoor}, our framework \textbf{ranks first} in the task of instance segmentation with limited annotations. The better performance has been achieved in small object-level semantic categories such as bookshelf and bed for ScanNet \cite{dai2017scannet} instance segmentation. For the overall performance, our method outperforms current art ReDAL by 1.5\% when there are 0.2\% labeled points. For the fine-grained object classes such as the bed and the bookshelf, our proposed approach also outperforms existing 3D WSL approaches such as PSD by a margin, which demonstrates the effectiveness of our approach.


As shown in Table~\ref{tablewslresults_sem_indoor} we have tested our methods in diverse circumstances with 20 labeled points to 200 labeled points respectively with the metrics of AP (Average Precision), AP 50\%, and AP 25\%. It turns out that our method ranks first in the instance segmentation with limited labeled points, outperforms current weakly supervised SOTA ContrastiveSceneContext by a great margin of at least 10\%, which demonstrates the superior performance of our framework. We have also tested our method in a complete unsupervised manner. We still reach AP of 46.8\% in the AP 50\% scenario for ScanNet, which demonstrates our hypothesis that the our instance segmentation results are accurate enough to instruct and provide supervision for the object detection.

As reported in Table \ref{tab_twist}, For the task of instance segmentation, we have also made comparisons with recent approach TWIST. Note that for a fair comparison, we have added contrastive losses proposed in PointContrast. It can be demonstrated that our proposed RM3D achieves comparable performance with TWIST in different labeling settings, and slightly overtakes TWIST by 1.1\% in the 1\% labeled setting, which further demonstrates the data-efficient learning capacity of our proposed RM3D.
\begin{table}[t]
\setlength{\abovecaptionskip}{-0cm}
\setlength{\belowcaptionskip}{-0cm}
\tiny
\begin{center}
\resizebox{\linewidth}{!}{\begin{tabular}{c|c|ccc|ccc}
\toprule[1pt]
\multirow{2}{*}{Settings} &\multirow{2}{*}{Method} & \multicolumn{3}{c|}{SemanticKITTI Semantic Seg.\%} & \multicolumn{3}{c}{Semantic3D Semantic Seg.\%} \cr & &mIoU\% & Sidewalk & Person & mIoU\% & buildings & cars 
\\
\midrule[1pt] 
\multirow{8}{*}{0.2\%}&WeakLabel-3DNet \textcolor{red}{(Our)} & \textcolor{red}{\textbf{50.9}}&70.8&55.8&\textcolor{red}{\textbf{74.8}}&91.4&80.2\\
&ReDAL\cite{wu2021redal}&48.7 & 68.2 &52.7 & 72.7&88.6&78.8\\

&PSD &47.6 & 67.8 & 51.1 &72.3& 87.4&75.8\\

&Xu. et al. \cite{liu2021one}&46.2&66.0&50.2&71.1&86.6 &74.6\\
&Viewpoint-Bottleneck \cite{luo2021pointly}&44.2&63.8&49.0&70.0&84.7&72.3\\

&ContrastiveSceneContext \cite{hou2021exploring}&42.5&61.6&47.8&66.6&78.7&69.8\\

&PointContrast \cite{xie2020pointcontrast}&41.1&59.3&45.8&61.8&74.7&66.5\\

&SparseConv Baseline \cite{graham20183d}&34.5 &53.9 &38.7&50.8&66.9&58.7\\
\midrule[1pt]
\multirow{7}{*}{One Pt.}&WeakLabel-3DNet \textcolor{red}{(Our)}&\textcolor{red}{\textbf{45.6}}&65.3&51.1&\textcolor{red}{\textbf{67.6}}&83.8&69.6\\

&ReDAL\cite{wu2021redal}&44.1&63.7&47.8&64.9&79.3&65.7\\

&PSD &42.5&62.1&46.2&63.5&78.1&61.4\\
&Xu. et al. \cite{liu2021one}&41.4&59.5&45.7&62.9&77.3&59.2\\
&Viewpoint-Bottleneck \cite{luo2021pointly}&39.7&56.9&43.1&60.3&74.2&57.3\\
&ContrastiveSceneContext \cite{hou2021exploring}&38.8&55.8&42.5&58.8&73.8&55.2\\
&PointContrast \cite{xie2020pointcontrast}&36.9&54.7&41.7&56.0&70.7&52.5\\
&SparseConv Baseline \cite{graham20183d} &30.1&49.3&44.6&49.9&65.1&47.1\\
\midrule[1pt]
\multirow{7}{*}{1.0\%}&WeakLabel-3DNet \textcolor{red}{(Our)}&\textcolor{red}{\textbf{53.7}}&75.6&59.2&\textcolor{red}{\textbf{75.3}}&93.8&81.3\\
&ReDAL \cite{wu2021redal}&51.9&73.4&55.1&74.5&92.9&80.5\\
& PSD  &51.1&73.1&54.5&73.9&91.2&79.7\\
&Xu. et al. \cite{liu2021one} &47.6&69.7&50.9&71.8&88.7&77.3\\
&Viewpoint-Bottleneck \cite{luo2021pointly}&46.9&67.3&48.5&69.7&85.9&76.9\\
&ContrastiveSceneContext \cite{hou2021exploring}&45.7&66.2&45.8&67.3&83.2&74.8\\

&PointContrast \cite{xie2020pointcontrast}&44.2&62.5&42.2&64.9&81.8&72.5\\

&SparseConv Baseline \cite{graham20183d}&38.6&55.6&35.7&55.8&74.9&63.5\\
\midrule[1pt]
100\%&Full-Supervised SparseConv \cite{graham20183d}&58.9&78.4&67.8&78.6&95.8&87.2\\
\bottomrule [1.5pt]
\end{tabular}}
\caption{The comparisons of the performance of our proposed method on \textbf{outdoor} benchmarks Semantic3D \cite{hackel2017semantic3d} and SemanticKITTI \cite{behley2019semantickitti} for the semantic segmentation task. The setting of One Pt. represents merely one labeled point for each instance class within the whole scene instead of small blocks (e.g. $1 \times 1 \times 1$ cubic meters) of Xu et al.. The diverse percentages of label ratios are tested. The backbone utilized for comparisons is unified as SparseConv.  All experimental results are three times on average.}
\label{tablewslresults_sem_outdoor}
\end{center}
\vspace{-5mm}
\end{table}


\begin{table}[t]
\setlength{\abovecaptionskip}{-0cm}
\setlength{\belowcaptionskip}{-0cm}
\tiny
\begin{center}
\resizebox{\linewidth}{!}{\begin{tabular}{c|c|ccc|ccc}
\toprule[1.5pt]
\multirow{2}{*}{Settings} &\multirow{2}{*}{Method} & \multicolumn{3}{c|}{KITTI Object Det. (Pedestrian)\%} & \multicolumn{3}{c}{Waymo Object Det. (Level-2 \underline{mAPH})\%} \cr & & Easy & Moderate & Hard & Vehicle & Pedestrian & Cyclist 
\\
\midrule[1pt]
\multirow{8}{*}{1.0\%}& WeakLabel-3DNet \textcolor{red}{(Our)} & \textcolor{red}{\textbf{52.3}}&\textcolor{red}{\textbf{45.3}}&\textcolor{red}{\textbf{42.7}}&\textcolor{red}{\textbf{63.3}}&\textcolor{red}{\textbf{50.4}}&\textcolor{red}{\textbf{56.2}}\\
&ReDAL&51.5 & 44.1 & 41.5 & 62.1&48.8 & 53.1\\

&PSD&50.6 & 43.0 & 40.1 & 61.2 & 47.9 & 52.3\\

&Xu. et al.& 48.6 & 41.9 & 39.3 & 60.3 &46.5 &50.1\\

&Viewpoint-Bottleneck&47.5& 40.8& 38.2 &59.3&45.6&49.4\\

&ContrastiveSceneContext &46.6&39.7&37.4&58.5&46.5&48.2\\

&PointContrast &45.7&38.5&36.6&57.7&45.3&47.3\\

&SparseConv Baseline&40.6 &33.2 & 31.3 &50.5&35.2&36.7\\

\midrule[1pt]
\multirow{7}{*}{One Box (0.5\%)}&WeakLabel-3DNet \textcolor{red}{(Our)}&\textcolor{red}{\textbf{45.6}}&\textcolor{red}{\textbf{37.6}}&\textcolor{red}{\textbf{34.3}}&\textcolor{red}{\textbf{57.8}}&\textcolor{red}{\textbf{45.6}}&\textcolor{red}{\textbf{51.6}}\\

&ReDAL&44.1&36.1&33.0&56.9&44.3&50.2\\

&PSD &42.5&36.2&32.2&55.7&43.4&48.7\\
&Xu. et al. &41.4&34.3&31.3&53.2&42.1&46.3\\
&Viewpoint-Bottleneck &39.7&33.7&30.6&52.2&41.0&45.0\\
&ContrastiveSceneContext &38.8&32.6&29.7&51.1&40.1&43.9\\
&PointContrast &36.9&30.7&26.4&48.7&38.2&42.2\\
&SparseConv Baseline  &32.7&26.6&22.9&41.9&30.5&35.1\\

\midrule[1pt]
\multirow{7}{*}{3.0\%}&WeakLabel-3DNet \textcolor{red}{(Our)}&\textcolor{red}{\textbf{54.1}}&\textcolor{red}{\textbf{47.8}}&\textcolor{red}{\textbf{43.8}}&\textcolor{red}{\textbf{65.6}}&\textcolor{red}{\textbf{53.6}}&\textcolor{red}{\textbf{59.3}}\\
&ReDAL &53.2&46.1&42.7&64.6&52.5&57.5\\
& PSD  &51.7&44.2&41.3&63.5&51.3&55.6\\
&Xu. et al.  &50.7&43.5&40.6&62.2&47.8&53.9\\
&Viewpoint-Bottleneck &49.6&42.5&39.5&60.8&46.9&52.6\\
&ContrastiveSceneContext &48.2&40.5&38.9&59.2&46.2&51.6\\

&PointContrast &46.6&39.8&37.8&57.9&46.9&50.1\\
&SparseConv Baseline &42.7&35.6&33.7&50.2&39.2&43.5\\
\midrule[1pt]
100\%&Full-Supervised SparseConv &57.7&49.5&47.0&68.9&62.6&65.6\\
\bottomrule [1.5pt]
\end{tabular}}
\caption{The comparisons of the performance of our proposed method on outdoor benchmarks KITTI and Waymo for the 3D object detection task in the 1\% and 3\% annotated bounding box case. The one box case denotes merely one randomly selected bounding box is annotated within ten point cloud scenes (0.5\% label). The backbone utilized for comparisons is unified as SparseConv. All experimental results are three times on average. The \underline{mAPH} denotes mean average precision weighted by heading. The Level-2 represents the level-2 detection difficulty, which is harder than level-1.}
\label{tablewslresults_det}
\end{center}
\vspace{-7mm}
\end{table}


\setlength{\tabcolsep}{3.0mm}{
\begin{table}[t]
\setlength{\abovecaptionskip}{-0.02cm}
\setlength{\belowcaptionskip}{-0.29888cm}
    \centering
    {
    \resizebox{\linewidth}{!}{
        \scalebox{0.78}{\begin{tabular}{l||l|l|l|l}
        \toprule [1.5pt]
           Label Ratio & 1\% & 5\% & 10\% & 20\%  \\
        \hline
                \textit{Sup-Only} & 17.6 & 47.0 & 58.9 & 63.0 \\
                
                \textit{PointContrast} & 20.3 & 48.5 & 59.5 & 63.6 \\
                
                CSC & 21.2 & 50.6 & 60.2 & 64.1 \\
                TWIST + CSC  & 31.1 & 58.2 & 53.7 & 67.8  \\
                RM3D + CSC (Ours) \cellcolor[RGB]{238,238,238}   &\textbf{32.2 \footnotesize{+1.1}} \cellcolor[RGB]{238,238,238}
        &
          \textbf{58.9 \footnotesize{+0.7}}   
          \cellcolor[RGB]{238,238,238}
        & 
          \textbf{53.9 \footnotesize{+0.2}} 
          \cellcolor[RGB]{238,238,238} 
        &
          \textbf{67.7 \footnotesize{-0.1}} 
          \cellcolor[RGB]{238,238,238}
        \\
          \bottomrule [1.5pt]
        \end{tabular}}}
    \caption{Data efficient indoor 3D instance segmentation average precision (AP\%) results with limited number of scene reconstructions compared with TWIST  on ScanNet with SparseConv as the backbone network. }
    \label{tab_twist}
    }
\vspace{-0.98mm} 
\end{table}}

\begin{table}[t!]
\begin{center}
\resizebox{\linewidth}{!}{\begin{tabular}{l|lll}
\toprule[1pt]
\multirow{2}{*}{Case No.} & \multicolumn{3}{c}{Diverse 3D scene understanding tasks} \cr & Sem. Seg. (mIoU\%)& Ins. Seg. (AP\%)& Obj. Det. (AP\%)\\
\midrule[1pt]                     
1. S3 $\rightarrow$ Sc (PFH $M_{\mbox{\textit{\tiny{{des}}}}}$)&59.7 ($\downarrow$ 5.5)  & 51.7 ($\downarrow$ 3.4)  & 32.9 ($\downarrow$ 3.0)    \\

2. S3 $\rightarrow$ Sc (w/o $M_{\mbox{\textit{\tiny{{des}}}}}$)& 48.2 ($\downarrow$ 17.0)  & 42.2 ($\downarrow$ 12.9)  & 24.2 ($\downarrow$ 11.7)  \\

3. S3 $\rightarrow$ Sc (w/o Oth.) &58.1 ($\downarrow$ 7.1)  & 48.6 ($\downarrow$ 6.5)  & 28.5 ($\downarrow$ 7.4)  \\

4. S3 $\rightarrow$ Sc (w/o $M_{\mbox{\textit{\tiny{{seg}}}}}$) &56.3 ($\downarrow$ 8.9)  & 46.7 ($\downarrow$ 8.4)  & 26.3 ($\downarrow$ 9.6)  \\

5. S3 $\rightarrow$ Sc (Predator $M_{\mbox{\textit{\tiny{{des}}}}}$) &58.4 ($\downarrow$ 7.8)  & 50.7 ($\downarrow$ 4.4)  & 31.3 ($\downarrow$ 4.6)  \\

6. Sc $\rightarrow$ S3 (PFH $M_{\mbox{\textit{\tiny{{des}}}}}$) & 60.3 ($\downarrow$ 5.9)  & 49.8 ($\downarrow$ 4.5) & 33.1 ($\downarrow$ 3.5) \\

7. Sc $\rightarrow$ S3 (w/o $M_{\mbox{\textit{\tiny{{des}}}}}$) & 49.3 ($\downarrow$ 16.9)  & 40.6 ($\downarrow$ 13.7) & 24.8 ($\downarrow$ 11.8) \\

8. Sc $\rightarrow$ S3 (w/o Oth.) & 58.2 ($\downarrow$ 8.0)  & 49.2 ($\downarrow$ 5.1) &  29.1 ($\downarrow$ 7.5)\\

9. Sc $\rightarrow$ S3 (w/o $M_{\mbox{\textit{\tiny{{seg}}}}}$) & 56.9 ($\downarrow$ 9.3)  & 48.6 ($\downarrow$ 5.7) &  27.5 ($\downarrow$ 9.1)\\

10. Sc $\rightarrow$ S3 (Predator $M_{\mbox{\textit{\tiny{{des}}}}}$) &57.3 ($\downarrow$ 8.9)  & 48.7 ($\downarrow$ 5.6)  & 31.6 ($\downarrow$ 5.0)  \\
\bottomrule
\end{tabular}}
\caption{\textit{RM3D} transfer learning performance for the task of the instance segmentation (Metric: AP@50\%) on the ScanNet/S3DIS validation Set. The $M_{\mbox{\textit{\tiny{{des}}}}}$ is selected as the similarity of our proposed adapted PFH-based 3D local descriptor. The annotation percentage is 0.2\% for the semantic and instance segmentation tasks and approximately 1\% for the object detection task. S3 $\rightarrow$ Sc denotes transferring trained model on S3DIS to ScanNet, and Sc $\rightarrow$ S3 denotes transferring trained model on ScanNet to S3DIS, respectively. }
\label{table_transfer_learning_i}
\end{center}
\vspace{-7mm}
\end{table}

\vspace{-3mm}

\vspace{-2mm}
\subsection{Results of WSL for Object Detection}
\vspace{-2mm}

We have also done experiments of object detection on KITTI and Waymo benchmarks with two annotated bounding box per ten scenes to keep fairness in comparisons, which means there are merely approximately 1\% labeled bounding boxes. Also, we have tested the circumstances of one box case, which denotes merely one randomly selected bounding box is annotated within ten point cloud scenes. It can be demonstrated that our framework outperforms other weakly supervised counterparts by a large margin. As shown in the quantitative experimental results in Table \ref{tablewslresults_det}, our proposed approach can also achieve superior performance for weakly supervised object detection compared with current weakly supervised learning approaches for the task of object detection.\\
\subsection{More Study on the Results of Transfer Learning}
\label{sub_trans_learning}

In this Subsection, we conduct more Studies on the results of transfer learning. As in demonstrated in our experimental results in Table \ref{table_transfer_learning_i}, the feature similarities of 3D local descriptor $M_{\mbox{\textit{\tiny{{des}}}}}$ plays a significant role in improving the generalization ability across different domains. For example, when transferring the model trained on S3DIS to ScanNet for the task of instance segmentation with 0.2\% label, the performance merely drops by 3.4\% and 4.5\% for S3DIS $\rightarrow$ ScanNet and ScanNet $\rightarrow$ S3DIS, respectively, demonstrating that our proposed model has good transfer learning capacity. And noticeably, for S3DIS $\rightarrow$ ScanNet, the performance drops ($\downarrow 12.9$) when removing the similarity of our adapted PFH-based 3D local descriptor $M_{\mbox{\textit{\tiny{{des}}}}}$ is much larger than the drop ($\downarrow8.4$) when removing the similarity in semantic predictions $M_{\mbox{\textit{\tiny{{Seg}}}}}$. It demonstrates that the our proposed adapted PFH-based 3D local feature descriptor-based similarity is of greater importance to the model generalization capacity compared with high-level semantic for the indoor case. Finally, when substituting the adapted PFH-based $M_{\mbox{\textit{\tiny{{des}}}}}$ with predator-based $M_{\mbox{\textit{\tiny{{des}}}}}$, the perofrmance will drop a little. It demonstrates the generalization capacity of our adapted PFH-based descriptor compared with learning-based ones.



In summary, according to our experimental results in Table \ref{table_transfer_learning_i}, the $M_{\mbox{\textit{\tiny{{des}}}}}$ is of great significance to the model generalization capacity for the tasks of semantic/instance segmentation and object detection. It is demonstrated that the generalization capacity of our adapted PFH-based 3D descriptor is better compared with learning-based descriptor Predator. It is also validated that the similarity in 3D local descriptor $M_{\mbox{\textit{\tiny{{des}}}}}$ is even more robust and significant to the overall performance than the learnt high-level semantic features similarities $M_{\mbox{\textit{\tiny{{Seg}}}}}$. Removing $M_{\mbox{\textit{\tiny{{des}}}}}$ results in a significant performance drop in transfer learning.  

\begin{table}[t!]
\begin{center}
\tiny
\resizebox{\linewidth}{!}{\begin{tabular}{l|cccccccc|cc}
\toprule[1pt]
 Cases  & Base & $A_{des}$ & $A_{n}$ & $M_{\mbox{\textit{\tiny{{des}}}}}$ & $M_{\mbox{\textit{\tiny{{Color}}}}}$ & $M_{\mbox{\textit{\tiny{{Scale}}}}}$ & $M_{\mbox{\textit{\tiny{{Seg}}}}}$ &WB & mIoU\% & AP@50\% \\
\midrule[1pt]  
No. 1 & \checkmark &  & \checkmark & \checkmark & \checkmark & \checkmark & \checkmark &\checkmark & 53.5 / 53.2 & 46.9 /  41.6 \\

No. 2 & \checkmark & \checkmark & & \checkmark & \checkmark & \checkmark  & \checkmark &\checkmark & 60.6 / 62.7 & 51.8 /  49.4 \\

No. 3 & \checkmark  & \checkmark  &  \checkmark &  & \checkmark & \checkmark & \checkmark &\checkmark & 51.1 / 55.6 & 45.0 /  40.8 \\

No. 4  & \checkmark  & \checkmark  &  \checkmark & \checkmark &  & \checkmark & \checkmark &\checkmark & 63.6 / 64.7 & 52.8 /  52.1 \\

No. 5  & \checkmark & \checkmark  & \checkmark  & \checkmark & \checkmark  & & \checkmark &\checkmark & 64.7 / 64.9 & 52.3 /  52.0 \\

No. 6 & \checkmark & \checkmark & \checkmark & \checkmark  & \checkmark & \checkmark & &\checkmark & 60.1 / 60.5 & 51.6 /  51.7 \\

No. 7 & \checkmark & \checkmark & \checkmark & \checkmark & \checkmark  & \checkmark & \checkmark &\checkmark & \textcolor{red}{\textbf{65.2}} / \textcolor{red}{\textbf{66.2}} & \textcolor{red}{\textbf{55.1}} /  \textcolor{red}{\textbf{54.3}}\\

No. 8 & \checkmark & \checkmark  & \checkmark &  & \checkmark  & \checkmark & \checkmark &\checkmark & 45.2 / 47.3 &  43.3 / 39.2\\

No. 9 & \checkmark & \checkmark  & \checkmark & \checkmark & \checkmark  & \checkmark & \checkmark & & 50.3 / 50.5 &  51.5 / 50.3\\
\bottomrule
\end{tabular}}
\caption{\textit{RM3D} ablation study of the influence of the proposed adapted PFH-based 3D local descriptor $M_{\mbox{\textit{\tiny{{des}}}}}$ and other diverse features on indoor ScanNet (Left Value) and S3DIS (Right Value) validation set, for the task of semantic segmentation (Metric: mIoU\%), and instance segmentation (Metric: AP@50\%) with the 0.2\% labeled points per scene case. The $M_{\mbox{\textit{\tiny{{des}}}}}$ is selected as the similarity of our proposed adapted PFH-based 3D local descriptor.} 
\label{table_ablation_fea_i}
\end{center}
\vspace{-5mm}
\end{table}
\vspace{-2mm}
\subsection{Ablation Studies}
\subsubsection{Ablation Studies on Influences of Various of Features}

\begin{table}[t!]
\begin{center}
\tiny
\resizebox{\linewidth}{!}{\begin{tabular}{l|cccccccc|c}
\toprule[1pt]
 Cases  & Base & $A_{des}$ & $A_{n}$ & $M_{\mbox{\textit{\tiny{{des}}}}}$ & $M_{\mbox{\textit{\tiny{{Color}}}}}$ & $M_{\mbox{\textit{\tiny{{Scale}}}}}$ & $M_{\mbox{\textit{\tiny{{Seg}}}}}$  &WB & \underline{mAPH}\%/AP\%  \\
\midrule[1pt]  
No. 1 & \checkmark &  & \checkmark & \checkmark & \checkmark & \checkmark & \checkmark & \checkmark & 48.6 / 37.7 \\

No. 2 & \checkmark & \checkmark & & \checkmark & \checkmark & \checkmark  & \checkmark & \checkmark & 57.2 / 39.5  \\

No. 3 & \checkmark  & \checkmark  &  \checkmark &  & \checkmark & \checkmark & \checkmark & \checkmark & 52.8 / 33.2 \\

No. 4  & \checkmark & \checkmark  & \checkmark  & \checkmark & \checkmark  & & \checkmark & \checkmark & 59.3 / 37.7\\

No. 5 & \checkmark & \checkmark & \checkmark & \checkmark  & \checkmark & \checkmark & & \checkmark & 57.5 / 36.5 \\

No. 6 & \checkmark & \checkmark & \checkmark & \checkmark & \checkmark  & \checkmark & \checkmark & \checkmark & \textcolor{red}{\textbf{63.3}} / \textcolor{red}{\textbf{42.7}} \\

No. 7 & \checkmark &  \checkmark & \checkmark &  & \checkmark  & \checkmark & \checkmark  & \checkmark & 42.2 / 31.6 \\

No. 8 & \checkmark & \checkmark & \checkmark & \checkmark & \checkmark  & \checkmark & \checkmark & & 60.8 / 40.2 \\
\bottomrule
\end{tabular}}
\caption{\textit{RM3D} ablation study of the influence of the proposed adapted PFH-based 3D local descriptor $M_{\mbox{\textit{\tiny{{des}}}}}$ and other diverse features on outdoor Waymo (Left Value) and KITTI (Right Value) validation set, for the task of Waymo object detection (Metric: mAPH\%), and and KITTI object detection (Metric: AP\%) with the 1\% label percentage. The $M_{\mbox{\textit{\tiny{{des}}}}}$ is selected as the similarity of our proposed adapted PFH-based 3D local descriptor.} 
\label{table_ablation_fea_o}
\end{center}
\vspace{-6mm}
\end{table}
We have done very detailed ablation study about our framework for various of features. And the conclusion is that we find that our adapted PFH-based feature is of great importance to the final segmentation performance. Also, our proposed adapted PFH-feature based 3D descriptor improve the generalization capacity across diverse domains, as is demonstrated in the Subsection \ref{sub_trans_learning}. \\
\textbf{Ablation Experiments} We did extensive ablation experiments of various features in the following settings. The final results are summarized in Table \ref{table_ablation_fea_i} and Table \ref{table_ablation_fea_o} for indoor and outdoor scene understanding, respectively. We have ablated network modules in all settings as follows. Take ScanNet instance segmentation at AP@50\% as examples: \textbf{Case 1}: Removing $A_{des}$ in the affinity calculation of the  $i_{th}$ and $j_{th}$ regions $A(R_i, R_j)$. This setting leads to a large drop of 8.2\% on AP. \textbf{Case 2}: Removing $A_{n}$ in the process of over-segmentation. The performance drops merely by 3.3\%. \textbf{Case 3}: Removing $M_{\mbox{\textit{\tiny{{des}}}}}$ in the region-level similarity prediction strategy. This setting leads to a significant drop of 10.1\% on AP. \textbf{Case 4}: Removing $M_{\mbox{\textit{\tiny{{Color}}}}}$ in the region-level similarity prediction strategy. The performance drops by 2.3\%. \textbf{Case 5}: Removing $M_{\mbox{\textit{\tiny{{Scale}}}}}$ in the region-level similarity prediction strategy. The performance drops by 2.8\%. \textbf{Case 6}:  Removing the learnt $M_{\mbox{\textit{\tiny{{Seg}}}}}$ in the region-level similarity prediction strategy. The performance drops by 3.5\%. \textbf{Case 7}:  Retaining all the features. It will result in the best segmentation performance both in semantic segmentation and instance segmentation. \textbf{Case 8}:  Removing $M_{\mbox{\textit{\tiny{{des}}}}}$ in the cluster-level similarity prediction strategy. This will results in the biggest performance drop of 11.8\% in AP@50\%.  \textbf{Case 9}:  Removing the weight balancing (denoted as WB in the Table \ref{table_ablation_fea_i} and \ref{table_ablation_fea_o}) strategy during training. This results in a slight performance drop of 3.6\% in AP@50\%.
\\
\begin{table}[t!]
\begin{center}
\tiny
\resizebox{\linewidth}{!}{\begin{tabular}{c|cccccc|cc}
\toprule[1pt]
\multirow{2}{*}{Case No.} & \multicolumn{3}{c}{ScanNet Semantic Segmentation\%} & \multicolumn{3}{c|}{ScanNet Instance Segmentation\%} & \multirow{2}{*}{\textbf{SS} mIoU\%} & \multirow{2}{*}{\textbf{IS} AP@50\%} \cr &\textbf{SS} $L^{js}_{Aug}$/$L^{mse}_{Aug}$ &\textbf{SS} $L_{\mbox{\textit{\tiny{WSL}}}}$ & Confidence Threshold $\gamma$ &\textbf{IS} $L^{js}_{Aug}$ /$L^{mse}_{Aug}$ & \textbf{IS} $L_{\mbox{\textit{\tiny{WSL}}}}$ & Confidence Threshold $\gamma$ \\
\midrule[1pt] 
No. 1 & \checkmark-1 &  \checkmark &\checkmark & N.A. & N.A. & N.A. &  \textcolor{red}{\textbf{65.2}} \textcolor{red}{\textbf{/ 66.2}} & N.A. \\
No. 2 & \checkmark-1 & &\checkmark & N.A. &  N.A. &N.A. & 43.7 / 45.1 &N.A.  \\
No. 3 &  & \checkmark & \checkmark&  N.A. & N.A. &N.A. &61.6 / 61.5 & N.A.   \\
No. 4 &\checkmark-1  & \checkmark & &  N.A. & N.A. &N.A. &59.8 / 60.3 & N.A.   \\
No. 5 &\checkmark-2  & \checkmark & \checkmark &  N.A. & N.A. &N.A. &62.8 / 62.3 & N.A.  \\
No. 6 & N.A. &  N.A. & N.A. & \checkmark-1 & \checkmark  & \checkmark &N.A. & \textcolor{red}{\textbf{55.1}}  \textcolor{red}{\textbf{/ 54.3}}\\
No. 7  & N.A.  & N.A.& N.A.  &  \checkmark-1 &  &  \checkmark  &N.A. & 41.7 / 40.5 \\
No. 8  & N.A. &  N.A. & N.A. &  & \checkmark &\checkmark & N.A. & 50.6 /  49.9 \\
No. 9  & N.A. &  N.A. & N.A. & \checkmark-1 & \checkmark &   &N.A. & 48.6 /  47.7 \\
No. 10  & N.A. &  N.A. & N.A. & \checkmark-2 & \checkmark & \checkmark  &N.A. & 52.3 /  51.9 \\
\bottomrule
\end{tabular}}
\caption{\textit{RM3D} ablation study of the network modules on ScanNet (Left Value) and S3DIS (Right Value) validation set, for the tasks of semantic segmentation (Metric: mIoU\%), and instance segmentation (Metric: AP@50\%) with 0.2\% labeled points per scene case. The $M_{\mbox{\textit{\tiny{{des}}}}}$ is selected as the similarity of our proposed adapted PFH-based 3D local descriptor.} 
\label{table_ablation_loss_i}
\end{center}
\vspace{-10mm}
\end{table}

For the outdoor ablations on the Waymo and KITTI as shown in Table \ref{table_ablation_fea_o} for the task of object detection, we also ablated the network in all settings. Take Waymo object detection as examples: \textbf{Case 1}: Removing $A_{des}$ in the affinity calculation of the $i_{th}$ and $j_{th}$ regions $A(R_i, R_j)$. This setting leads to a large drop of 14.7\% on \underline{mAPH}\%. \textbf{Case 2}: Removing $A_{n}$ in the process of over-segmentation. The AP@50\% drops merely by 6.1\%. \textbf{Case 3}: Removing $M_{\mbox{\textit{\tiny{{des}}}}}$ in the cluster-level similarity prediction strategy. This setting leads to a significant drop of 10.5\% on \underline{mAPH}\%.  \textbf{Case 4}: Removing $M_{\mbox{\textit{\tiny{{Scale}}}}}$ in the cluster-level similarity prediction strategy. The performance drops by 4.0\%. \textbf{Case 5}:  Removing the learnt $M_{\mbox{\textit{\tiny{{Seg}}}}}$ in the cluster-level similarity prediction strategy. The performance drops by 5.8\%. \textbf{Case 6}:  Retaining all the features. It will result in the best object detection performance both for the Waymo and KITTI object detection. \textbf{Case 7}:  Removing $M_{\mbox{\textit{\tiny{{des}}}}}$ in the cluster-level similarity prediction strategy simultaneously. This will results in the biggest performance drop of 21.1\% in \underline{mAPH}\%. \textbf{Case 8}:  Removing the weight balancing (WB) strategy during training. This results in a slight performance drop of 2.5\% in in \underline{mAPH}\%.\\
\textbf{Ablation Analyses} The experimental results show that our proposed adapted PFH-based 3D local feature descriptor plays a significant role in the performance of the whole network. The adapted PFH-based feature descriptor is of great significance in the cluster-level similarity prediction strategy. It is more powerful than the other low-level features such as the color, the scale, and the normal feature for the task of semantic/instance segmentation and for the task of object detection. It demonstrates that PFH is still a very powerful local feature descriptor in the deep learning area. And when combining the low-level geometric PFH feature with the learnt high-level semantic features, the best performance can be achieved. It can be interpreted by the hypothesis that the neural network will inevitably overlook some low level geometric details, which can be exactly captured by the PFH-based local descriptors. The future work includes designing a fusion mechanism for the adapted PFH feature and learnt high-level semantic feature in a better way for better fulfilling the task of 3D scene understanding. Further, it can be demonstrated that the weight balancing strategy can also boost the weakly supervised 3D scene understanding performance on multiple tasks.

\begin{table}[t!]
\begin{center}
\tiny
\resizebox{\linewidth}{!}{\begin{tabular}{c|cccc|c}
\toprule[1pt]
\multirow{2}{*}{Case No.} & \multicolumn{4}{c|}{KITTI and Waymo Object Detection\%}  & \multirow{2}{*}{mAPH\%/AP\%} \cr & $L^{js}_{Aug}$ & $L^{mse}_{Aug}$ & $L_{\mbox{\textit{\tiny{WSL}}}}$ & Confidence Threshold $\gamma$ &\\
\midrule[1pt] 
No. 1 & \checkmark-1 & &  \checkmark &\checkmark &  \textcolor{red}{\textbf{63.3}} / \textcolor{red}{\textbf{42.7}} \\
No. 2 & \checkmark-1 & & &\checkmark & 51.2 / 33.8  \\
No. 3 & & & \checkmark& \checkmark& 59.5 / 41.3 \\
No. 4 &\checkmark-1 & & \checkmark&  &60.8 / 40.5 \\
No. 5 &  & \checkmark-2 & \checkmark&\checkmark & 59.8 / 40.8 \\
\bottomrule
\end{tabular}}
\caption{\textit{RM3D} ablation study of the network modules on Waymo with the evaluation metric of mAPH$\%$ (Left Value) and KITTI with the evaluation metric of AP$\%$ (Right Value) validation set, for the tasks of object detection with 0.2\% labeled points per scene case. The $\checkmark$-1 denotes using $L^{js}_{Aug}$ for the network optimizations, and $\checkmark$-2 denotes using $L^{mse}_{Aug}$ for the network optimizations. The $M_{\mbox{\textit{\tiny{{des}}}}}$ is selected as the similarity of our proposed adapted PFH-based 3D local descriptor.}
\label{table_ablation_loss_det_o}
\end{center}
\vspace{-0.2mm}
\end{table}

\subsubsection{Ablation Studies on Different Network Optimization Functions for Semantic/Instance Segmentation}
We have also done a very detailed ablation study of the optimization loss functions of our framework, as shown in Table \ref{table_ablation_loss_i} for indoor benchmarks, and Table \ref{table_ablation_loss_det_o} for outdoor benchmarks. Take ScanNet semantic segmentation and instance segmentation at AP@50\% as examples, as shown in Table \ref{table_ablation_loss_i}, we have ablated the network modules in all six settings as follows: \textbf{Case 1}: Retain both the loss function for the data augmentation submodule $L^{js}_{Aug}$, and the loss for weakly supervised learning $L_{\mbox{\textit{\tiny{WSL}}}}$. as well as the confidence threshold $\gamma$ in region merging. This setting will result in the best semantic segmentation performance of 65.2\% mIoU for ScanNet. \textbf{Case 2}: Removing the $L_{\mbox{\textit{\tiny{WSL}}}}$ for semantic segmentation, it means that we have discarded our proposed adapted PFH-based region expansion strategies, and the segmentation losses are substituted by the cross entropy loss supervised by the weak labels. The results is that the semantic segmentation performance on ScanNet drops significantly from 65.2\% to 43.7\%, which is a 21.5\% performance drop. It demonstrates the effectiveness of our proposed cluster-level similarity prediction strategy and learning-based region merging submodule. \textbf{Case 3}: We have removed the loss of the data augmentation submodule $L^{js}_{Aug}$. It results in a marginal 3.6\% performance drop in the task of ScanNet semantic segmentation. \textbf{Case 4}: Removing the confidence threshold $\gamma$. It results in a 5.4\% performance drop in the task of semantic segmentation, which demonstrates that the confidence regularization is significant to retain the regions with high confidence to provide high-quality pseudo labels, \textbf{Case 5}: We have retained all losses and substitute the $L^{js}_{Aug}$ in the data augmentation loss with the mean square error-based loss $L^{mse}_{Aug}$. It yields a 2.4\% performance drop in the task of ScanNet semantic segmentation. Compared with case 1 with $L^{js}_{Aug}$, It demonstrates our proposed JS divergence-based loss $L^{js}_{Aug}$ is of great significance to the overall performance, which can be explained by the fact that $L^{js}_{Aug}$ can help the network learn an optimized distribution and reduce overfitting compared with the mean square error-based loss. It further validates that JS divergence-based loss is significant to the performance and is better than the mean square error-based loss $L^{mse}_{Aug}$. \textbf{Cases 6, 7, 8, 9, 10} are for instance segmentation. \textbf{Case 6}: The full network will result in the best performance in 55.1\% AP@50\% for the ScanNet instance segmentation task. \textbf{Case 7}: Removing the $L_{\mbox{\textit{\tiny{WSL}}}}$ will result in a significant drop of 13.4\% at AP@50\%. It proves that our proposed cluster-level similarity prediction strategy and learning-based region merging submodule are also of great significance to the performance of the task of instance segmentation. \textbf{Case 8}: Removing $L^{js}_{Aug}$ merely results in a slight drop of 4.5\% on ScanNet instance segmentation. \textbf{Case 9}: Removing the confidence threshold $\gamma$ leads to a 6.5\% performance drop. \textbf{Case 10}: Retaining all losses and substituting the $L^{js}_{Aug}$ in the data augmentation loss with the mean square error-based loss $L^{mse}_{Aug}$ will cause a drop of 2.8\%.

\begin{table}[t]
\begin{center}
\resizebox{\linewidth}{!}{\begin{tabular}{c|ccc|c}
\toprule[1pt]
\multirow{2}{*}{Case No.} & \multicolumn{3}{c|}{ScanNet 3D Object Detection\%} & \multirow{2}{*}{AP@50\%} \cr & $L_{Seg,2}$ & $L_{Dice}$ & $L_{cls}$ \\
\midrule[1pt] 
No. 1 & \checkmark & \checkmark & \checkmark  &  \textcolor{red}{\textbf{35.9}} \\

No. 2 &  & \checkmark & \checkmark & 21.3 \\

No. 3 & \checkmark  &   &  \checkmark &    29.2 \\

No. 4 & \checkmark  & \checkmark &  & 27.5  \\

\bottomrule
\end{tabular}}
\caption{\textit{RM3D} ablation study of the network optimization loss functions for the task of 3D object detection on ScanNet validation set (Metric: AP@50\%) with two annotated bounding boxes (approximately 1\%) per ten scenes.} 
\label{table_ablation_detect_i}
\end{center}
\vspace{-6.8mm}
\end{table}

For the outdoor case, the results are reported in Table \ref{table_ablation_loss_det_o}. We ablated the network in the outdoor case in all settings as follows: \textbf{Case 1}: Retain all network loss modules. Our performance achieves 63.3\% for the task of object detection for the class vehicle in the Waymo object detection.  \textbf{Case 2}: Removing the $L_{\mbox{\textit{\tiny{WSL}}}}$ for object detection, the performance drops by 12.1\%, which is a significant drop. It demonstrates the effectiveness of our proposed cluster-level similarity prediction strategy and learning-based region merging submodule. \textbf{Case 3}: We remove the loss of the data augmentation submodule $L^{js}_{Aug}$. It results in a performance drop of 3.8\%. \textbf{Case 4}: We remove the confidence threshold $\gamma$.This results in a 2.5\% performance drop in the task of semantic segmentation. It demonstrates that confidence regularization is significant to retain the regions with high confidence to provide high-quality pseudo labels during region merging. \textbf{Case 5}:  We have retained all losses and substituted the $L^{js}_{Aug}$ in the data augmentation loss with the mean square error-based loss $L^{mse}_{Aug}$. It results in a 3.5\% performance drop for object detection, which demonstrates the effectiveness of our proposed JS divergence-based data augmentation loss in the task of 3D scene understanding.  In summary, the ablation studies for indoor scene understanding reported in Table \ref{table_ablation_loss_i} and outdoor 3D scene understanding in Table \ref{table_ablation_loss_det_o} verify the effectiveness of the proposed modules of our \textit{RM3D}.

\vspace{-2mm}
\begin{table}[t]
\begin{center}
\resizebox{\linewidth}{!}{\begin{tabular}{c|ccc|c}
\toprule[1pt]
\multirow{2}{*}{Case No.} & \multicolumn{3}{c|}{KITTI/Waymo 3D Obj. Det.\%} & \multirow{2}{*}{\underline{mAPH}\%/AP\%} \cr & $L_{Seg,2}$ & $L_{Dice}$ & $L_{cls}$ \\
\midrule[1pt] 
No. 1 & \checkmark & \checkmark & \checkmark  &  \textcolor{red}{\textbf{63.3/}}\textcolor{red}{\textbf{42.7}} \\

No. 2 &  & \checkmark & \checkmark & 43.1/27.9 \\

No. 3 & \checkmark  &   &  \checkmark &  58.5/36.3 \\

No. 4 & \checkmark  & \checkmark &  & 52.2/34.8 \\
\bottomrule
\end{tabular}}
\caption{\textit{RM3D} ablation study of the network optimization loss functions for the task of the object detection on KITTI (Metric: AP\%, left value) and Waymo  validation set (Metric: \underline{mAPH}\%, right value) with two annotated bounding boxes (approximately 1\%) per ten scenes.}
\label{table_ablation_detect_o}
\end{center}
\vspace{-3mm}
\end{table}

\subsubsection{Ablation Experiments on Different Network Optimization Functions for Object Detection}

We have done a wide range of ablation experiments based on the proposed optimization loss functions. We have also done experiments of ScanNet object detection with 1\% annotations with the evaluation metric of AP@50\%, and the related results are shown in Table~\ref{table_ablation_detect_i}. For the task of object detection, we have ablated the network modules in all the four settings as follows: \textbf{Case 1}: We have tested with the full proposed network for the task of object detection. Our proposed network can realize the performance of 35.9\% at AP@50\%, which establishes current SOTAs in 3D object detection as shown in the Table \ref{table_ablation_detect_i}. We also test the experimental results in the public benchmark, and our method realized the SOTAs performance. \textbf{Case 2}: We have removed the loss $L_{Seg,2}$ from the instance segmentation branch. The 3D object detection performance drops greatly from 35.9\% to 21.3\% by a percent of 14.6\%. \textbf{Case 3}: We have removed the Dice loss $L_{Dice}$, the segmentation performance drops by 6.7\%, which demonstrates the effectiveness of the Dice loss in our design. It can be explained by the fact that Dice loss can tackle well with the unbalanced classes, thus improving the performance. \textbf{Case 4}: We have also removed the loss $L_{cls}$, which is the loss for predicting the presence of an object within a scene or not. The object detection performance also drops by percent of 8.4\%, which demonstrates the effectiveness of or proposed the scene-level object classes loss in the weakly supervised indoor 3D object detection.     

We have also done ablation studies for the task of object detection on KITTI and Waymo validation set. The results are reported in Table \ref{table_ablation_detect_o}. Take Waymo object detection as examples, we have ablated the network modules in all the four settings as follows: \textbf{Case 1}: We have tested with the full proposed network for the task of object detection. Our proposed network can realize the performance of 63.3\%. \textbf{Case 2}: We have removed the loss $L_{Seg,2}$ from the instance segmentation branch. The 3D object detection performance drops greatly from 63.3\% to 43.1\% by a percent of 20.2\%. \textbf{Case 3}: We have removed the Dice loss $L_{Dice}$, the segmentation performance drops by 4.8\%, which demonstrates the effectiveness of the Dice loss in our design. It can be explained by the fact that Dice loss can tackle well with the unbalanced classes, thus improving the performance. \textbf{Case 4}: We have also removed the loss $L_{cls}$, which is the loss for predicting the presence of an object within a scene or not. The object detection performance also drops by percent of 11.1\%, which demonstrates the effectiveness of or proposed the scene-level object classes loss in the weakly supervised outdoor 3D object detection. 
\section{Conclusion}
In conclusion, we have conducted a comprehensive algorithm comparison of the traditional and learning-based 3D descriptors in the tasks of over-segmentation and downstream weakly-supervised 3D scene understanding.  Based on that, we have proposed a general benchmark framework for weakly supervised point clouds understanding which has superior performance for the three most significant semantic understanding tasks including 3D semantic segmentation, 3D instance segmentation, and 3D object detection. Our proposed framework can be integrated seamlessly with diverse 3D descriptors to realize downstream 3D scene understanding tasks. The proposed network learns to merge over-divided regions based on the local geometric property similarities and the learnt feature similarities. Network modules are proposed to fully investigate the relations among semantics/geometric relationships within a scene, thus, high-quality pseudo labels can be generated from weak labels to provide a better segmentation supervision. The effectiveness of our approach is verified across diverse large-scale real-scene point clouds understanding benchmarks under various test circumstances with excellent rotational robustness.

\vspace{-3mm}
\bibliographystyle{spmpsci}
{\footnotesize
\bibliography{shortstrings,vggroup,cvww_template,mybib}
}

\end{document}